\documentclass{article}
\usepackage[T1]{fontenc}
\usepackage{graphicx} 
\usepackage{amsmath, amssymb, graphicx}
\usepackage[a4paper, total={6in, 8in}]{geometry}
\usepackage{natbib}
\usepackage{float}
\usepackage{tabularx}
\usepackage{multirow}
\usepackage[table]{xcolor} 
\usepackage{colortbl} 
\usepackage{booktabs}
\usepackage{array}
\usepackage{caption}
\usepackage{subcaption}
\usepackage{xcolor}
\usepackage{longtable}
\usepackage{tikz}
\usepackage{float} 
\usepackage{enumitem}
\usetikzlibrary{positioning}
\usepackage{adjustbox}
\usepackage[colorlinks=true, allcolors=blue]{hyperref}
\tikzstyle{startstop} = [rectangle, rounded corners, minimum width=4cm, minimum height=1.2cm, text centered, draw=black]
\tikzstyle{process} = [rectangle, minimum width=4cm, minimum height=1.2cm, text centered, draw=black]
\tikzstyle{textbox} = [rectangle, minimum width=4cm, text width=4.5cm, minimum height=4cm, text centered, draw=black] 
\tikzstyle{arrow} = [thick,->,>=stealth]

\title{Combating the Bullwhip Effect in Rival Online Food Delivery Platforms Using Deep Learning}
\author{Tisha Ghosh}
\date{}

\begin{document}

\maketitle

\begin{abstract}
The wastage of perishable items has caused significant health and economic crises, leading to uncertainty in the business environment and fluctuating customer demand. In modern times, this issue has intensified with the rise of online food delivery services, where frequent and unpredictable orders contribute to inefficiencies in supply chain management. These fluctuations contribute to the bullwhip effect, a major supply chain issue that results in stockouts, excess inventory, and inefficiencies. Accurate demand forecasting and predictive analysis help stabilize inventory, optimize supplier orders, and reduce resource wastage.
This paper proposes a Third-Party Logistics (3PL) supply chain model involving key stakeholders— restaurants, online food apps, and customers—alongside a deep learning-based demand forecasting model using a two-phase Long Short-Term Memory (LSTM) network. Phase 1 (Intra-day demand forecasting) captures short-term variations within a day, while Phase 2 (Daily demand forecasting) predicts overall daily demand. A two-year historical dataset (January 2023 – January 2025) with daily sales data from Swiggy and Zomato is used to capture nonlinear demand trends through discrete event simulation. The grid search method is employed to select optimal LSTM hyperparameters, ensuring precise forecasting. The proposed method is evaluated and compared using Root Mean Square Error (RMSE), Mean Absolute Error (MAE), and R² score, along with optimized training time, among which R² serves as the primary measure of forecasting accuracy. Phase 1 achieves an R² score of 0.69 for Zomato and 0.71 for Swiggy with a training time of 12 minutes, while phase 2 shows substantial improvement, with R² values of 0.88 (Zomato) and 0.90 (Swiggy), achieving a training time of 8 minutes, demonstrating strong predictive performance in demand forecasting. To mitigate demand fluctuations, restaurant inventory is dynamically managed using the newsvendor model, adjusted based on forecasted demand. The proposed framework significantly reduces the bullwhip effect, enhancing forecasting accuracy and supply chain efficiency. For Phase 1, the overall supply chain instability decreases from 2.61 to 0.96, while for Phase 2, it reduces from 2.19 to 0.80, demonstrating the model’s effectiveness in minimizing food waste and maintaining optimal inventory levels of restaurants.

\end{abstract}


Discrete event simulation, Bullwhip effect.

\section{Introduction}
Supply Chain Management (SCM) has been a critical area of research for a long time, with a particular focus on the management of perishable goods \citep{blackburn2009supply}. The food supply chain, in particular, faces numerous challenges, including stockouts, overstocking, ingredient waste, bottlenecks, and the bullwhip effect \citep{li2014sustainable, folkerts1997challenges}. In today’s fast-paced world, the younger generation has embraced the convenience of online food delivery apps, making them an essential part of daily life. However, this growing reliance on food delivery services has led to unpredictable demand patterns, resulting in frequent stockouts and inventory fluctuations. To compensate, restaurants and suppliers often overstock to avoid shortages, which results in excessive wastage of perishable items. This issue is a manifestation of the bullwhip effect, where minor changes in consumer demand lead to increasingly significant fluctuations in orders and stock levels as the supply chain moves upstream \citep{lee1997bullwhip}. Usually, it is caused by factors like inaccurate demand forecasting, order batching, pricing changes, and information flow delays. To ensure smoother supply chain operations, managing the bullwhip effect requires improved supply chain coordination, demand forecasting techniques and accurate information sharing. Demand is one of the most crucial data points that must be communicated among supply chain participants \citep{abolghasemi2020demand}, and effective demand forecasting play a crucial role in addressing these challenges \citep{haberleitner2010implementation, kumar2020big}, ensuring customer satisfaction while preventing supply chain disruptions.

Traditional forecasting models, such as ARIMA (AutoRegressive Integrated Moving Average), Holt-Winters moving average, and exponential smoothing, have been widely used for demand prediction \citep{moroff2021machine, abbasimehr2020optimized}. However, these methods are primarily linear and often fail to accurately capture real-world demand patterns, which tend to be non-linear. To address these limitations, non-linear statistical models like ARCH (Autoregressive Conditional Heteroskedasticity) and GARCH (Generalized Autoregressive Conditional Heteroskedasticity) have been introduced \citep{khashei2011novel}. While these models offer improvements over purely linear approaches, they still struggle to fully capture complex non-linear dependencies in demand fluctuations. 

As a result, companies are seeking more advanced methods to closely align with customers' needs.
In response to these challenges, businesses have increasingly turned to machine learning (ML) techniques, which have demonstrated superior predictive capabilities across various industries \citep{min2010artificial}. ML-based forecasting models have been successfully applied to demand prediction in sectors such as furniture \citep{abbasimehr2020optimized}, energy \citep{szul2020neural}, cash flow at ATMs \citep{catal2015improvement}, tourism \citep{shahrabi2013developing}, and natural gas supply chains \citep{potovcnik2019comparison}. Some of the most widely used ML techniques for time series forecasting include Artificial Neural Networks (ANNs), Support Vector Machines (SVMs), K-Nearest Neighbors (KNNs), and Adaptive Neuro-Fuzzy Inference Systems (ANFIS)\citep{abbasimehr2020optimized}.

Among these, Recurrent Neural Networks (RNNs) have gained significant attention due to their ability to recognize sequential dependencies in time-series data and predict future trends based on past observations \citep{amirkolaii2017demand}. Recurrent Neural Networks (RNNs) are widely applied across various areas of the supply chain. For instance, \citep{aktepe2021demand} demonstrated the superiority of RNNs in forecasting spare parts demand. Similarly, \cite{rahman2018predicting} showcased their effectiveness in medium- and long-term electricity consumption predictions. Additionally, \cite{potovcnik2019comparison} compared RNNs with linear regression and extreme machine learning algorithms for forecasting natural gas demand, incorporating factors such as historical temperatures and time-based variables, including holidays and other seasonal events. Their findings highlighted the superiority of RNNs. However, research also indicates that RNNs are not always the optimal choice for supply chain applications \citep{amirkolaii2017demand}. Their limitations stem from short memory retention and the vanishing gradient problem, which complicates training and reduces their effectiveness in certain scenarios. Unlike traditional ANNs, RNNs incorporate feedback loops, allowing information to persist across time steps, making them particularly effective for time-dependent forecasting tasks. However, RNNs suffer from short-term memory limitations \citep{hochreiter1997long} and the vanishing gradient problem, which reduces their effectiveness for long-term forecasting applications. To overcome these drawbacks, Long Short-Term Memory (LSTM) networks were developed. LSTMs are specifically designed to retain relevant information over extended periods by distinguishing between short-term and long-term dependencies. This characteristic makes them highly effective for time series forecasting, as they can learn patterns in demand fluctuations over both short and long durations \citep{wu2018remaining, siami2019comparative}. After successfully capturing demand patterns, companies stabilize their inventory.

In this study, a two-phase multilayer LSTM model is developed and optimized using the grid search technique, which systematically explores different hyperparameter configurations to enhance forecasting accuracy. The proposed model is applied to third-party logistics (3PL) providers in the online food delivery industry who forecast demand, such as Uber Eats, DoorDash, and Zomato. They act as competitive demand aggressors (CDA) by connecting business-to-consumer (B2C) with business-to-business (B2B) restaurant suppliers in highly active Indian cities where online food delivery is most prevalent, utilizing synthetic sales data for 2 years on two competitive platforms, such as Zomato and Swiggy, to validate their performance. This research contributes to the field of supply chain in several ways. First, a synthetic data set is simulated with relevant columns, replicating real-world online food sales data from highly active Indian cities where Zomato and Swiggy operate. Second, it develops and fine-tunes a two-phase multilayer LSTM model using the GridSearch method to achieve optimal forecasting precision- Phase 1 (Intra-day demand forecasting) captures short-term fluctuations within a single day and Phase 2 (Daily demand forecasting) predicts overall daily demand trends for improved inventory planning. It develops and fine-tunes a multi-layer LSTM model using the Grid Search method to achieve optimal forecasting precision. Third, inventory replenishment is optimized based on dynamically forecasted demand fluctuations, preventing both overstocking and shortages. Fourth, the reduction in the bullwhip effect is measured by analyzing demand fluctuations and the stabilization of inventory replenishment for restaurants using online food delivery platforms, both at the individual platform level and overall within the supply chain.

The remainder of this study presents a detailed analysis of the proposed approach. A review of related research on demand forecasting and supply chain management is provided in Section $2$. Section $3$ outlines the problem description, including the simulation approach, mathematical formulation of the LSTM model, and strategies for reducing the bullwhip effect. This is followed by an explanation of the methodology used to develop the two phase LSTM model, optimize its hyperparameters using Grid Search, and preprocess the dataset in Section $4$. In Section $5$, the numerical setup and process are discussed, including all numerical values used. Section $6$ presents the results and discussion, analyzing the impact of the proposed model on reducing the bullwhip effect and improving inventory stability. Finally, in Section $7$, the study concludes with a summary of key findings and managerial implications, outlining actionable steps for companies to implement the proposed model.

\section{Literature review}
In this section, the literature in four different but related streams of our proposed model is reviewed, namely, $(i)$ bullwhip effect in SCM, $(ii)$ data simulation for bullwhip effect analysis $(iii)$ online food delivery applications and challenges, and $(iv)$ demand forecasting techniques. 

\subsection{Bullwhip effect in SCM}
Over the past few decades, the bullwhip effect on product orders has become a widely studied topic among researchers and practitioners. Early studies attempted to demonstrate the existence of the bullwhip effect on product orders and identify its underlying causes \citep{forrester1997industrial, forrester2012industrial}. The bullwhip effect is a phenomenon in which minor variations in customer demand lead to progressively larger fluctuations in orders as they propagate upstream in the supply chain \citep{lee1997bullwhip}. Currently, most research focuses on quantifying the bullwhip effect and exploring strategies to mitigate its impact. \cite{lee2000value} provided a formal definition of the bullwhip effect on product orders and systematically analyzed four main causes: demand signal processing, order batching, price fluctuations, and rationing. Additionally, they proposed several countermeasures, including reducing frequent updates of demand forecasts, breaking order batching, stabilizing prices, and eliminating gaming behavior in shortage situations \citep{lee1997bullwhip, lee1997information}. Several studies have emphasized the role of demand forecasting in influencing the bullwhip effect \citep{sodhi2012researchers}. Researchers have also identified the bullwhip effect across various industries, including retail \citep{yao2008inventory, ramanathan2014performance}, automotive \citep{ bai2020supply}, healthcare and pharmaceuticals \citep{costantino2015spc}, electronics and semiconductor manufacturing \citep{lee1997information, chen2012bullwhip} and food supply chains \citep{yang2021behavioural}. Over time, various strategies have been developed to mitigate the bullwhip effect. Early studies identified key causes and solutions, such as information sharing \citep{lee1997bullwhip}, emphasizing accurate demand data to reduce uncertainty. Order batching reduction emerged as another approach, as \cite{lee1997bullwhip} advocated just-in-time replenishment, later supported by \cite{peng2014manage} for Haier Group’s case. Price stabilization was recommended by \cite{lee1997bullwhip} to minimize demand spikes. In the 2000s, demand forecasting improvement was explored by \cite{chen2000impact}, showing how better forecasting reduces variability, while lead time reduction was highlighted by \cite{lee1997bullwhip} for its role in faster demand response. In the 2010s, vendor-managed inventory (VMI) gained attraction when \cite{ravichandran2008managing} proved its effectiveness in consumer durables and also in semiconductors \citep{diaz2022simulated}. Agile and lean practices were also studied, with \cite{peng2014manage} demonstrating their success at Haier. Additionally, modern forecasting models improved with \cite{li2024mitigating} by applying robust control theory to minimize order fluctuations. These studies illustrate a progressive shift from fundamental supply chain coordination mechanisms to sophisticated, data-driven, and technology-enhanced solutions for bullwhip effect reduction.

\subsection{Data simulation for bullwhip effect analysis}
Simulation techniques play a crucial role in studying and mitigating the bullwhip effect by replicating real-world supply chain dynamics. Various studies have explored simulation techniques to mitigate the bullwhip effect. System dynamics simulations were used by \cite{sterman1989modeling} and \cite{dejonckheere2003measuring} to model demand amplification and optimize replenishment policies. Agent-based modeling (ABM) was applied by \cite{nair2011supply} to analyze decentralized decision-making impacts on demand variability. Monte Carlo simulations and bootstrapping have been used for sales data generation, with \cite{sterman1989modeling} applying them to retail forecasting. GANs were introduced by \cite{tkachuk2025consumer} to simulate realistic sales patterns. Time-series simulations often use ARIMA models \citep{zhang2003time} or stochastic distributions like Poisson and Gaussian \citep{syntetos2005accuracy}. While \cite{chatfield2004bullwhip} used DES to measure demand amplification under different inventory policies. \cite{wangphanich2010analysis} integrated stochastic demand patterns into DES to assess forecasting impacts. In our study, we adopt DES with stochastic distributions to model demand and lead time variability, allowing for a more realistic assessment.

\subsection{Online food delivery applications and Challenges}
Online food delivery (OFD) applications have revolutionized the food industry, improving convenience but also introducing significant supply chain challenges. \cite{chopra2001strategy} highlighted the importance of real-time tracking to enhance supply chain efficiency, while \cite{hwang2020effects} explored drone-based delivery as a potential future solution. Demand fluctuations create inefficiencies, with sudden order surges leading to stockouts and delays, as noted by \cite{suali2024role}. However, through machine learning forecasting accuracy has been improved \citep{aamer2020data}. Logistics and last-mile delivery constraints, such as traffic congestion and workforce shortages, impact delivery speed, but predictive analytics and dynamic routing provide optimization strategies \citep{chu2023data}. Inventory management is another concern, with perishable food wastage resulting from inaccurate predictions \citep{osman2023perishable}. Environmental concerns, including excessive packaging waste and carbon emissions, have led to sustainability initiatives such as eco-friendly packaging and electric vehicles \citep{maheshwari2021internet}, with regulatory policies further promoting green delivery practices \citep{pal2023optimizing}. Technological advancements like blockchain and IoT are also transforming supply chains, with blockchain improving transparency and traceability \citep{zhang2022iot} and IoT-enabled temperature monitoring ensuring food safety in transit \citep{mohsin2017iot}. These studies highlight the complexities of OFD supply chains and the role of innovation in addressing them. This problem is particularly pronounced on competitive food delivery platforms, where third-party logistics (3PL) providers serve as competitive demand aggregators (CDAs), connecting B2C customers with B2B restaurant suppliers. Unlike traditional supply chains, food delivery networks face highly volatile demand patterns, short fulfillment cycles, and perishable inventory, necessitating more advanced forecasting and inventory optimization techniques.

\subsection{Demand forecasting techniques}
Effective demand forecasting and inventory smoothing are critical for maintaining supply chain efficiency, reducing costs, and improving customer satisfaction. Various techniques, ranging from traditional statistical models to advanced machine learning approaches, have been explored to optimize inventory management. To improve forecasting accuracy and mitigate demand variability, researchers have explored more sophisticated forecasting methods such as ARIMAX models, which integrate external factors into forecasting \citep{assad2023comparing, babai2022demand}. More recently, machine learning-based forecasting techniques have been explored for their potential to reduce the bullwhip effect. \cite{feizabadi2022machine} showed that machine learning models outperform traditional statistical approaches while \cite{paruthipattu2021demand} compared AI-based techniques such as Random Forest, XGBoost, and LSTM, concluding that they provide superior accuracy in demand prediction. \cite{sariciouglu2024analyzing} further investigated LSTM networks, demonstrating their effectiveness in mitigating supply chain fluctuations. Other studies, such as those by \cite{rezki2024deep}, explored hybrid models integrating deep learning with statistical forecasting, while \cite{kiuchi2024recurrent} examined reinforcement learning algorithms, highlighting their adaptability to dynamic demand patterns \citep{borucka2023seasonal, zhu2021demand}.

The above literature review demonstrates effective supply chain management, and the evolution of techniques has mitigated the issues starting from the very beginning of traditional supply chain models. However, managing the current perishable food supply chains, which are highly volatile and competitive in the modern era, is a big challenge. Therefore, as motivations to develop the current model, the following key research gaps are highlighted:
\begin{itemize}
    \item Many studies using simulation-based approaches primarily focus on long-term demand trends while overlooking short-term variations, seasonal changes, and event-driven spikes (e.g., holidays, promotions, or sudden market shifts).
    \item Advanced forecasting techniques, particularly deep learning-based models such as LSTM and reinforcement learning, often require high computational power and careful hyperparameter tuning.
    \item Limited integration of AI-driven forecasting models with adaptive inventory control systems tailored for perishable goods.
\end{itemize}
This paper addresses these challenges by simulating sales data with short-term fluctuations, seasonal variations, and event-driven demand spikes. Unlike studies focusing only on long-term trends, our LSTM-based model captures both short- and long-term patterns through optimized hyperparameter tuning, enhancing forecasting accuracy and mitigating the bullwhip effect. Table \ref{tab:literature} summarizes key recent papers in each of these streams, highlighting their contributions and relevance to our study
\begin{table}[htbp]
\centering
\footnotesize{
\begin{tabular}{|>{\raggedright}m{5cm}|m{1.8cm}|m{2.2cm}|m{1.8cm}|m{1.8cm}|}
\hline
{Author(s)} & {Simulation} & {Deep Learning\newline Forecasting} & {Bullwhip Effects} & {Online Food\newline Delivery}\\ \hline
\cite{viloria2019demand} & & \checkmark & &\\ \hline
\cite{alabdulkarim2020minimizing} & \checkmark & & \checkmark &\\ \hline
\citep{aamer2020data} & & \checkmark & & \checkmark\\ \hline
\cite{chiadamrong2021meta} & \checkmark & & \checkmark &\\ \hline
\cite{ramirez2021neural} & & \checkmark & \checkmark &\\ \hline
\cite{yang2022supply} & \checkmark & & \checkmark &\\ \hline
\cite{tjen2023demand} & & \checkmark & \checkmark &\\ \hline
\cite{udenio2023exponential} & \checkmark & & \checkmark &\\ \hline
\cite{rezaeefard2024present} & & \checkmark & \checkmark &\\ \hline
\cite{kleinemolen2024inventory} & \checkmark & & \checkmark &\\ \hline
\cite{chen2024inventory} & \checkmark & & \checkmark & \\ \hline
This study &\checkmark & \checkmark & \checkmark & \checkmark\\ \hline
\end{tabular}}
\caption{Comparison of our study with the existing literature}
\label{tab:literature}
\end{table}

Our study addresses these gaps by proposing a novel AI-enhanced framework that integrates predictive demand analytics with adaptive inventory management strategies to enhance the stability and efficiency of online food delivery supply chains while mitigating bullwhip effects.

\section{Problem description}
This supply chain model follows an online food delivery system, integrating key stakeholders: restaurants (as suppliers and service providers), third-party logistics (3PLs), customers, and the online platform as the central coordinator.

\begin{figure}[ht]
    \centering
    \includegraphics[width=0.5\linewidth]{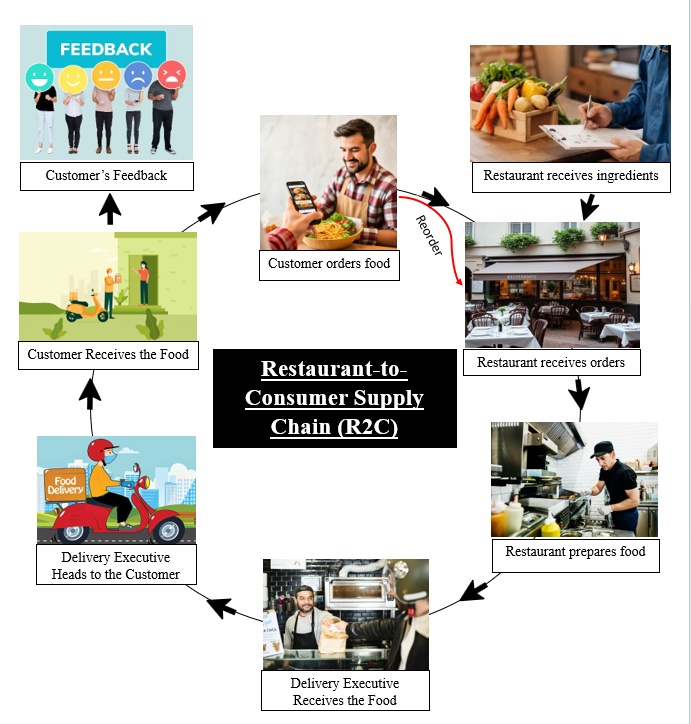}
    \caption{Supply chain model for online food delivery platforms representing the interactions between customers, Third-party logistics providers (3PLs), and restaurant suppliers}
    \label{fig:enter-supply}
\end{figure}

The process begins with restaurants managing inventory and sourcing ingredients. When a customer places an order through the online app, the platform directs it to the restaurant. The restaurant prepares the meal and hands it over to a 3PL delivery executive, who ensures timely delivery. Once the customer receives the order, they provide feedback via the app, contributing to service improvements for all stakeholders, as shown in Figure \ref{fig:enter-supply}.

\newpage
This section outlines the supply chain model developed to address the bullwhip effect in competitive online food delivery platforms. The proposed framework integrates three key components: discrete event simulation (DES), long short-term memory (LSTM) for demand forecasting, and bullwhip effect calculation.
\subsection{Discrete event simulation (DES)}
Discrete event simulation (DES) is a modeling technique used to analyze complex systems where events occur at distinct points in time. It is widely applied in supply chain and logistics management to simulate real-world operations and evaluate system performance under varying conditions. In this study, DES is employed to replicate and analyze the operations of two competitive online food delivery platforms, accounting for demand fluctuations, dynamic pricing, lead times, and external disruptions such as weather, holidays, and special events. DES is particularly useful in this context as it captures dynamic interactions between multiple stakeholders, examines how fluctuations in demand impact order fulfillment and inventory levels, and evaluates the bullwhip effect. Additionally, it provides valuable insights for optimizing pricing strategies, delivery efficiency, and inventory management, ultimately helping to develop strategies that mitigate demand uncertainties and improve overall supply chain resilience.
\subsubsection{Simulation framework}
    The simulation models the demand for Swiggy and Zomato, two leading competating online food delivery apps, based on patterns from two established Kaggle datasets \footnote{\href{https://www.kaggle.com/datasets/bhanupratapbiswas/food-delivery-time-prediction-case-study/data}{\url{https://www.kaggle.com/datasets/bhanupratapbiswas/food-delivery-time-prediction-case-study/data}}, \\ \href{https://www.kaggle.com/datasets/cbhavik/swiggyzomato-order-information}{\url{https://www.kaggle.com/datasets/cbhavik/swiggyzomato-order-information}}}. It operates on a time-based framework from January 2023 to January 2025, where events are triggered at five specific timesteps daily: morning, noon, evening, night, and midnight. The simulation runs continuously, generating demand for a 24-hour cycle using SimPy.

\subsubsection{Key assumptions}

This simulation model is built upon several theoretical assumptions that govern the demand patterns, pricing mechanisms, and supply chain dynamics.
\begin{enumerate}
    \item [a)] Demand functions
    \begin{itemize}
        \item \textit{Long-term trends and seasonality:} The demand for food delivery services follows both a long-term trend (growth or decline) and seasonal variations across five distinct periods.
    \item \textit{Cyclical demand patterns:} The fluctuation of demand throughout the day is modeled using a sinusoidal function \footnote{\href{https://mtimpa.weebly.com/uploads/1/9/3/5/19359645/chap06.2.pdf}{\url{https://mtimpa.weebly.com/uploads/1/9/3/5/19359645/chap06.2.pdf}}}
:
\begin{equation}
C(t) = B_{\text{phase}} \sin\left(\frac{\pi}{P} t + c\right) + h
\end{equation}

where \( B_{\text{phase}} \) represents the phase amplitude, \( P \) controls periodicity, \( c \) is a phase shift constant, and \( h \) is a baseline adjustment for both swiggy and zomato.
    \item \textit{Seasonality factors (S(t)):} Various external factors contribute to fluctuations in demand, including
    \begin{itemize}
        \item \textit{Time of day:} Demand varies significantly across different hours, peaking during meal times.\footnote{\href{https://tfresource.org/topics/Model_Validation_and_Reasonableness_Checking_Time_Of_Day.html}{\url{https://tfresource.org/topics/Model_Validation_and_Reasonableness_Checking_Time_Of_Day.html}}}

        \item \textit{Weather conditions:} Extreme or mild weather conditions influence customer ordering behavior.\footnote{\href{https://climatechange.chicago.gov/climate-impacts/climate-impacts-agriculture-and-food-supply}{\url{https://climatechange.chicago.gov/climate-impacts/climate-impacts-agriculture-and-food-supply}}}

        \item \textit{Holidays and special events:} Demand experiences spikes during major holidays and cultural festivals.\footnote{\href{https://www.sonwillogistics.com/unwrapping-the-impact-of-the-holiday-season-on-the-food-industry/}{\url{https://www.sonwillogistics.com/unwrapping-the-impact-of-the-holiday-season-on-the-food-industry/}}}

    \end{itemize}
     \item \textit{Stochastic demand across platforms:} The model accounts for two distinct platforms, each exhibiting its own stochastic demand curve with random noise, following a normal distribution to reflect real-world variability.
    \end{itemize}
    \item[b)] {Market and customer segmentation:} Different customer groups exhibit distinct ordering patterns, influenced by factors such as demographics, purchasing power, and regional preferences.
    \item[c)] {Dynamic pricing mechanism:} Pricing is dynamically adjusted based on market trends, platform-specific pricing strategies, and demand-supply equilibrium.
    \item[d)] {Order fulfillment and logistics:}  The order fulfillment process is modeled using an \textit{M/G/$\infty$} queuing system, where order arrival rate follows an exponential distribution (M), capturing the randomness of incoming orders. Service time (G) accounts for both food preparation and delivery duration, where preparation time is assumed to follow a lognormal distribution and delivery time follows a uniform distribution with infinite servers.
    \item[e)] {Inventory uncertainty:} The availability of food ingredients fluctuates based on demand patterns, supplier restocking schedules, and external disruptions.
\end{enumerate}

\subsubsection{Mathematical Formulation}
The demand for food delivery platforms is influenced by price, lead time, and external factors. The proposed mixed demand model, as stated in \citep{chopra2007supply}, captures these dynamics as given below.
\begin{itemize}
\item[a)]\textit{Demand Function}\\
The demand functions of the two food delivery platforms are given below
\begin{align}
    D_Z(t_i) &= \big[ \alpha_Z(t) - (\beta_Z(t_i) \cdot P_Z(t_i)) - (\gamma(t) \cdot L_Z(t_i)) \notag \\
    &\quad + \tau(t) (P_S(t_i) - P_Z(t_i)) - \delta(t) (L_Z(t_i) - L_S(t_i)) + \epsilon(t) \big] \times F(t)
\end{align}

\begin{align}
    D_S(t_i) &= \big[ \alpha_S(t) - (\beta_S(t_i) \cdot P_S(t_i)) - (\gamma(t) \cdot L_S(t_i)) \notag \\
    &\quad + \tau(t) (P_Z(t_i) - P_S(t_i)) - \delta(t) (L_S(t_i) - L_Z(t_i)) + \epsilon(t) \big] \times F(t)
\end{align}

where Z refers Zomato and S refers Swiggy and
\begin{itemize}
    \item t = The number of days the dataset is simulating
    \item $t_i$= Each day taking 5 times (i= morning, noon, evening, night, midnight)
    \item $D_Z(t_i), D_S(t_i)$ = Demand for each platform
    \item $\alpha_Z(t), \alpha_S(t)$ = Baseline demands
    \item $\beta_Z(t_i), \beta_S(t_i)$ = Price sensitivity coefficients
    \item $P_Z(t_i), P_S(t_i)$ = Platform prices for different food
    \item $\tau(t)$, $\gamma(t), \delta(t)$ = Price competition intensity factor, lead time sensitivity coefficient, lead time competition intensity factor
    \item $L_Z(t_i), L_S(t_i)$ = Platform lead times
    \item $\epsilon(t)$ = Random Noise follows Normal$(\mu(t),\sigma^2(t))$
    \item T(t)= Time of the day
    \item E(t)= Event or holiday effect
    \item $\beta_{customer}$= Customer segmentation
    \item F(t) = Cumulative effect of seasonal and external factors\\
    \[
     F(t) = T(t)\times W(t) \times E(t) \times \beta_{customer}
    \]
\end{itemize}
Taking the expectation on both sides
\begin{equation}
\begin{aligned}
    E[D_Z(t_i)] &= \big[ \alpha_Z(t) - \beta_Z(t_i) P_Z(t_i) - \gamma(t) E[L_Z(t_i)] \\
    &\quad + \tau(t) (P_S(t_i) - P_Z(t_i)) - \delta(t) (E[L_Z(t_i)] - E[L_S(t_i)]) \\
    &\quad + E[\epsilon(t)] \big] \times F(t)
\end{aligned}
\end{equation}
\begin{equation}
\begin{aligned}
    E[D_S(t_i)] &= \big[ \alpha_S(t) - \beta_S(t_i) P_S(t_i) - \gamma(t) E[L_S(t_i)] \\
    &\quad + \tau(t) (P_Z(t_i) - P_S(t_i)) - \delta(t) (E[L_S(t_i)] - E[L_Z(t_i)]) \\
    &\quad + E[\epsilon(t)] \big] \times F(t)
\end{aligned}
\end{equation}
and the variances are
\begin{equation}
    \text{Var}(D_Z(t_i)) = (\gamma^2(t) + \delta^2(t)) \text{Var}(L_Z(t_i)) + \delta^2(t) \text{Var}(L_S(t_i)) - 2\delta(t) \text{Cov}(L_Z(t_i), L_S(t_i)) + \sigma^2(t)
\end{equation}
\begin{equation}
    \text{Var}(D_S(t_i)) = (\gamma^2(t) + \delta^2(t)) \text{Var}(L_S(t_i)) + \delta^2(t) \text{Var}(L_Z(t_i)) - 2\delta(t) \text{Cov}(L_Z(t_i), L_S(t_i)) + \sigma^2(t)
\end{equation}

Here \( L_Z(t_i) \) and \( L_S(t_i) \) are uncorrelated as they are lead time of two different platforms.

\item[b)] \textit{Lead time function} \\
The lead time for an order on a platform (e.g., zomato, swiggy) is defined as:
\begin{equation}
    L_k(t_i) = T_{\text{s},k} + T_{\text{r},k}
\end{equation}

where:

- \( T_{\text{s},k} \) = Time when the order delivery is fulfilled to the customer.\\
- \( T_{\text{r},k} \) = Time when the restaurant accepts the customer’s order.\\
- k= swiggy, zomato

We are using the M/G/\(\infty\) model as it effectively represents the dynamics of food delivery systems. These systems experience random order arrivals and variable service times while ensuring that each order is assigned to a delivery executive without delay.  This model captures the following components
\begin{itemize}
    \item {Markovian or poisson process (M):} The orders arrive randomly over time, and the inter-arrival times \( T_{\text{r},i} \) follow an exponential distribution with rate $\lambda_i$
\begin{equation}
    P(T_{\text{r},i} > t) = e^{-\lambda_i t}, \quad t \geq 0
\end{equation}
    \item {General Service Time Distribution (G):} 
The total service time for an order consists of two components: the food preparation time and the delivery time. This can be expressed as
\begin{equation}
    T_{\text{s},i} = T_{p,i} + T_{d,i}
\end{equation}

Where (\(T_{p,i}\)) represents the food preparation time and (\(T_{d,i}\)) represents the delivery time. The food preparation time(\(T_{p,i}\)) is assumed to follow a lognormal distribution. This choice is based on the positive skewness typically observed in cooking times, which arise due to variability in preparation speed, human involvement, and different cooking methods \citep{thecharmsofs2001og}. The lognormal distribution is defined as
\begin{equation}
    T_{p,i} \sim \text{Lognormal}(\mu_i, \sigma_i)
\end{equation}

Where $\mu_i$, $\sigma_i$ are the parameters of the lognormal distribution, representing the mean and standard deviation of the logarithm of the preparation times, respectively. The probability density function (PDF) for the lognormal distribution is:
\begin{equation}
    f_{T_{p,i}}(t) = \frac{1}{t\sigma_i\sqrt{2\pi}} \exp\left(-\frac{(\ln t - \mu_i)^2}{2\sigma_i^2}\right), \quad t > 0
\end{equation}

The cumulative distribution function (CDF) for this distribution is given by
\begin{equation}
    F_{T_p}(t) = \frac{1}{2} + \frac{1}{2} \operatorname{erf}\left(\frac{\ln t - \mu_i}{\sigma_i \sqrt{2}}\right)
\end{equation}

Next, the delivery time ($T_d$) is assumed to be influenced by the distance 
between the restaurant and the customer. Given that delivery times are bounded within a certain range, we model $T_d$ using a uniform distribution. This reflects the fact that the delivery time can vary depending on the distance but is limited by geographical constraints and operational conditions\citep{christopher2022logistics}. Thus, we assume
\begin{equation}
    T_{d,i} \sim \text{Uniform}(a_i, b_i)
\end{equation}

For a {uniform distribution}, the probability density function (PDF) is

\begin{equation}
    f_{T_{d,i}}(t) = \frac{1}{b_i - a_i}, \quad a_i \leq t \leq b_i
\end{equation}

where \( a_i \) and \( b_i \) are the lower and upper bounds of the delivery time, respectively.

The cumulative distribution function (CDF) for the uniform distribution is

\begin{equation}
    F_{T_{d,i}}(t) = \frac{t - a_i}{b_i - a_i}, \quad a_i \leq t \leq b_i
\end{equation}
    \item {$\infty$ (Infinite Servers):} customers wait if all servers are busy in traditional queues. However, in food delivery, each order is immediately assigned to a delivery executive as soon as it is ready. Since there is no limit on the number of delivery executives available, we assume an infinite number of servers. This prevents congestion and ensures smooth operation.
\end{itemize}

\item[c)] \textit{Price function}\\
A real-time dynamic pricing strategy is formulated as
\begin{equation}
    P_i =(P_b \times G_i) + F_i + B_i +(R_i \times D_i) + S_i
\end{equation}

{Where}
- i= swiggy, zomato\\
- \( P_b \) = Base prices for for food as per Indian chart \\
- \( G_i \) = GST multiplier  \\
- \( F_i \) = Platform fees for zomato and swiggy  \\
- \( B_i \) = Base delivery fees for zomato and swiggy  \\
- \( R_i \) = Per km delivery rate \\
- \( D_i \) = Distance from restaurant to customer house\\
- \[
S_i =
\left\{
\begin{array}{ll}
\text{Small\_order\_fee}, & \text{if } P_b \leq \text{threshold} \\
0, & \text{otherwise}
\end{array}
\right.
\]
This pricing formula is obtained directly from the platform and reflects the real-time dynamic pricing model used by food delivery aggregators.
\item[d)] \textit{Restaurant inventory ordering} \\
Perishable food ingredients have a limited shelf life, making inventory management a critical challenge. To minimize the risk of overstocking while ensuring sufficient supply, the newsvendor model provides an effective approach \citep{hadley1963analysis}. In our case, demand originates from two independent sources, such as swiggy and zomato.
This study employs a two-phase inventory management approach using the newsvendor model, integrating both historical demand-based inventory calculation and LSTM forecast-driven optimization. The methodology is divided into two cases
\begin{itemize}
    \item Case 1: Inventory Calculation Using 5-Time-Per-Day Demand Predictions
    \item Case 2: Inventory Calculation Using Daily Average Demand Predictions
\end{itemize}

In case 1 at each time step \( t_i \) (where i= represents morning, noon, evening, night, and midnight), the total demand is given by

\begin{equation}
    D(t_i) = D_Z(t_i) + D_S(t_i)
\end{equation}

Since demand follows a distribution, the expected total demand (mean) is

\begin{equation}
    \mu_{\text{5-time, total}} = E[D_Z(t_i)] + E[D_S(t_i)]
\end{equation}

The total standard deviation is

\begin{equation}
    \sigma_{\text{5-time,total}} = \sqrt{\sigma^2_{Z(t_i)} + \sigma^2_{S(t_i)} + 2\rho \sigma_{Z(t_i)} \sigma_{S(t_i)}}
\end{equation}

where:
\begin{itemize}
    \item \( \sigma_{Z(t_i)} \) and \( \sigma_{S(t_i)} \) are the standard deviations of Zomato and Swiggy demands for each time of a day.
    \item \( \rho \) is the correlation coefficient between $D_Z(t_i)$ and $D_S(t_i)$.
\end{itemize}

The inventory level is determined as \cite{hadley1963analysis}

\begin{equation}
    Q_{\text{5-time}}^* = \mu_{\text{5-time,total}} + z \cdot \sigma_{\text{5-time,total}}
\end{equation}
 where \( z \) is the Z score of the standard normal distribution.

In case 2, inventory is updated once per day, using the daily average demand instead of frequent intra-day updates.
\begin{equation}
    \mu_{\text{daily}} = \frac{1}{5} \sum_{t=1}^{5} \mu_{\text{5-time,total}}
\end{equation}

The daily standard deviation is

\begin{equation}
    \sigma_{\text{daily}} = \frac{1}{5} \sum_{t=1}^{5} \sigma_{\text{5-time,total}}
\end{equation}

Thus, the daily inventory level follows

\begin{equation}
    Q_{\text{daily}}^* = \mu_{\text{daily}} + z \cdot \sigma_{\text{daily}}
\end{equation}
After performing the two-phase LSTM approach, the inventory is updated as follows:

For intra-day demand predictions using Phase 1 LSTM, the inventory is updated at each time step \( t_i \):

\begin{equation}
    Q_{\text{5-time,lstm}}^* = \mu_{\text{5-time,total lstm}} + z \cdot \sigma_{\text{5-time, total lstm}}
\end{equation}

where the demand at each time \( t_i \) is updated using LSTM predictions:

\begin{equation}
    D(t_i) \rightarrow D(t_i, \text{lstm})
\end{equation}

For the daily-level demand prediction using Phase 2 LSTM, the inventory is updated as:

\begin{equation}
    Q_{\text{daily,lstm}}^* = \mu_{\text{daily lstm}} + z \cdot \sigma_{\text{daily lstm}}
\end{equation}

where the daily demand prediction is computed as

\begin{equation}
    D(t, \text{lstm}) = \frac{1}{5} \sum_{i=1}^{5} D(t_i, \text{lstm})
\end{equation}

This two-phase approach ensures that inventory is dynamically adjusted based on intra-day fluctuations while maintaining a stable daily-level forecast.

\end{itemize}

\subsection{Long short-term Mmemory (LSTM) model architecture}
 LSTM networks are a specialized type of RNN designed to capture both long- and short-term dependencies. Since its introduction by Hochreiter and Schmidhuber in 1997\citep{hochreiter1997long}, they have been refined and widely adopted. While traditional RNNs can theoretically model long-term dependencies, their effectiveness is limited due to issues like vanishing gradients.
\begin{figure}[H]
    \centering
    \begin{subfigure}[t]{0.5\textwidth}
        \centering
        \includegraphics[width=\linewidth]{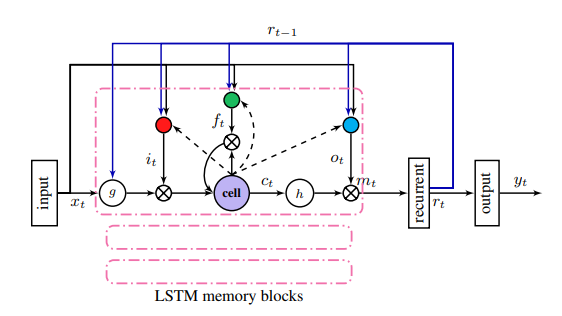}
        \caption{LSTM RNN architecture. A single memory block is shown for clarity.}
        \label{fig:enter-lstm}
    \end{subfigure}%
    \hfill
    \begin{subfigure}[t]{0.45\textwidth}
        \centering
        \includegraphics[width=\linewidth]{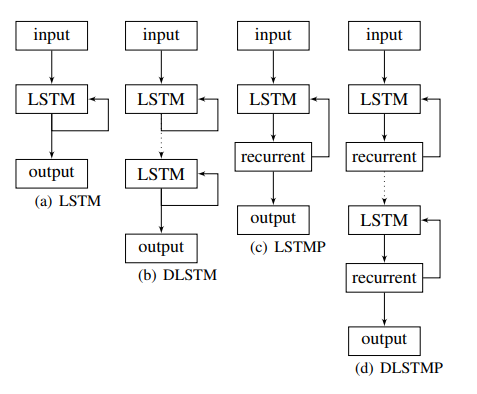}
        \caption{LSTM RNN architectures.}
        \label{fig:enter-lstm2}
    \end{subfigure}
    \caption{Illustration of LSTM architectures.}
\end{figure}

LSTMs address this limitation by incorporating memory cells and a gating mechanism that selectively retains or discards information.
These gates help regulate the flow of historical information, ensuring relevant past data is preserved while redundant details are discarded.
Figures \ref{fig:enter-lstm} \& \ref{fig:enter-lstm2} depict the description of the LSTM memory
block to capture sequential dependencies in demand forecasting follows that of \cite{sak2014long}.
The specific steps of the LSTM algorithm are as follows:
    \begin{align}
        i_t &= \sigma(W_i h_{t-1} + W_x X_t + b_i) \quad &\text{(Input Gate)} \nonumber \\
        f_t &= \sigma(W_f h_{t-1} + W_x X_t + b_f) \quad &\text{(Forget Gate)} \nonumber\\
        o_t &= \sigma(W_o h_{t-1} + W_x X_t + b_o) \quad &\text{(Output Gate)} \nonumber\\
        \tilde{c}_t &= \tanh(W_c h_{t-1} + W_x X_t + b_c) \quad &\text{(Candidate Cell State)} \nonumber \\
        c_t &= f_t \cdot c_{t-1} + i_t \cdot \tilde{c}_t \quad &\text{(Final Cell State Update)} \nonumber\\
        h_t &= o_t \cdot \tanh(c_t) \quad &\text{(Hidden State Update)} \nonumber
    \end{align}
where,
    \begin{itemize}
        \item \( i_t, f_t, o_t \) are the input, forget, and output gate activations.
        \item \( h_t \) represents the hidden state at time \( t \).
        \item \( c_t \) is the cell state at time \( t \), while \( \tilde{c}_t \) is the candidate cell state.
        \item \( W_i, W_f, W_o, W_c, W_x \) are weight matrices governing transformations.
        \item \( b_i, b_f, b_o, b_c \) are bias terms.
    \end{itemize}

\noindent The hyperbolic tangent activation function is given by

    \begin{equation}
    \tanh(x) = \frac{e^x - e^{-x}}{e^x + e^{-x}}
    \end{equation}
Here, we use a two-phase LSTM approach.

\begin{itemize}
    \item \textit{Phase 1 (Intraday forecasting):} The model takes a sequence of five time steps within a single day and predicts the next day's five time steps. This phase helps in capturing short-term variations in demand that occur within a single day.
    \item \textit{Phase 2 (Daily forecasting):} The model takes sequences spanning six days and predicts the demand for the next day. This phase is essential for capturing long-term temporal patterns and trends across multiple days.
\end{itemize}

In both phases, the LSTM processes sequential data, leveraging its memory cell to store and update information effectively. The intraday phase ensures that short-term fluctuations are well modeled, while the daily phase allows the model to account for broader trends and cyclical patterns in demand.

\subsection{Measure of bullwhip effect}
The bullwhip effect refers to the amplification of demand variability as it moves upstream in the supply chain, leading to inefficiencies such as excessive inventory holding, stockouts, and unpredictable order fluctuations. This effect is quantified using the bullwhip coefficient 
B, defined as \citep{rezaeefard2024present}
\[
B=\frac{\sigma^2_I}{\sigma^2_d}
\]
where \( \sigma_I^2 \) represents the variance of inventory present at the restaurant and \( \sigma_d^2 \) represents the variance of customer demand (either for an individual platform or total across platforms). 

If \( B > 1 \), demand amplification is occurring, leading to inefficiencies such as excessive inventory, stockouts, and unstable order patterns.
If \( B < 1 \), demand is well-regulated, meaning order variability is lower than demand variability, resulting in a stable supply chain.

In the context of a restaurant using Swiggy and Zomato as selling platforms, minimizing the bullwhip effect is crucial to maintaining a stable and efficient supply chain. The bullwhip coefficients for each platform are given by
   \[
   B_{\text{Zomato}} = \frac{\sigma_{I,\text{Zomato}}^2}{\sigma_{d,\text{Zomato}}^2}, \quad
   B_{\text{Swiggy}} = \frac{\sigma_{I,\text{Swiggy}}^2}{\sigma_{d,\text{Swiggy}}^2}
   \]
The overall bullwhip coefficient across both platforms is
   \[
   B_{\text{Overall}} = \frac{\sigma_{I,\text{Total}}^2}{\sigma_{d,\text{Total}}^2}
   \]

\noindent The objective is to minimize $B$ at both individual and overall levels (such that $B<1$), so that demand amplification is reduced and a stable, efficient order fulfillment system is maintained.
\section{Proposed methodology}
Designing a deep learning model for time-series forecasting in an online food delivery application requires a structured approach to handle demand fluctuations efficiently. Selecting the optimal forecasting model involves leveraging modern deep learning algorithms to capture complex demand patterns.

\begin{figure}
    \centering
    \includegraphics[width=0.8\linewidth]{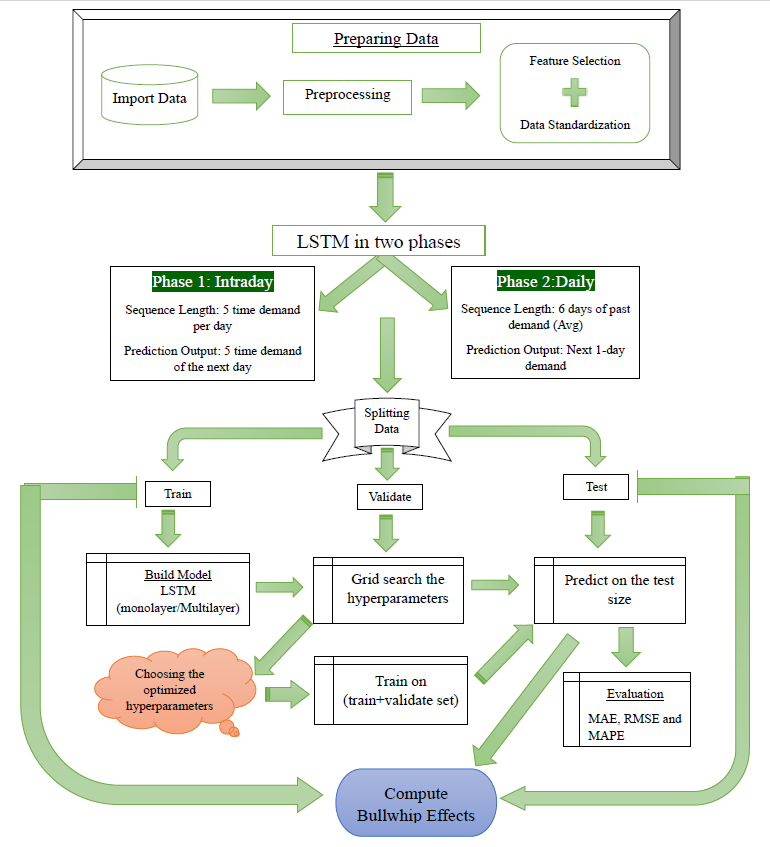}
    \caption{Flowchart of the proposed method}
    \label{fig:enter-flowchart}
\end{figure}

The proposed methodology for food order demand forecasting is illustrated in Figure \ref{fig:enter-flowchart}), outlining the key steps involved in predicting order volumes, optimizing inventory, and enhancing delivery efficiency.
\subsection{Data preprocessing}
The data used in this study represents the demand history of an online food delivery service, specifically for Zomato and Swiggy. The time series dataset spans from January 2023 to January 2025, capturing fluctuations in order volumes over this period. Before delving into the analysis of the time series, it is essential to properly refine and structure the raw data. In this study, feature selection, category encoding, standardization, and additional feature selection were performed as key preprocessing steps to ensure the dataset is well-prepared for accurate demand forecasting.

\subsubsection{Categorical encoding}
Machine learning models require numerical inputs, but categorical variables contain valuable information. Encoding these variables prevents data loss while ensuring they are properly interpreted by the model. In our model, we applied one-hot encoding to categorical features such as weather conditions (sunny, rainy, cloudy, etc.) and food categories (fast food, desserts, beverages, etc.), as these are nominal variables with no inherent order.

For one-hot encoding, each categorical variable is transformed into a vector of binary values

\begin{equation}
X_{\text{encoded}} = \{x_1, x_2, \dots, x_n\}, \quad x_i \in \{0,1\}
\end{equation}
For event importance (high, medium, low), we used ordinal encoding since it follows a meaningful order. We assigned numerical values as follows

\[
X_{\text{event}} =
\begin{cases} 
2, & \text{if high} \\
1, & \text{if medium} \\
0, & \text{if low}
\end{cases}
\]
\subsubsection{Feature selection}  

Selecting relevant features improves model performance by reducing noise, lowering computational complexity, and preventing overfitting. In our model, the dependent variable (\( Y \)) represents demand, while the independent variables (\( X \)) were chosen based on their impact on order forecasting for both Swiggy and Zomato

\begin{equation}
X = \{P_i(t), L_i(t), \Delta_i(t), E(t)\}
\end{equation}

where \( P_i(t) \) represents the price of the food item, \( L_i(t) \) denotes the lead time, \(\Delta_i(t)\) denotes the delivery distance (i=Swiggy, Zomato), and \( E(t) \) captures external factors, including the public holidays, special events, Weather conditions. These features were selected separately for Swiggy and Zomato to account for platform-specific variations while ensuring an optimal demand forecasting model.
\subsubsection{Standardization}  

The goal of standardization is to scale the values of the source dataset to have a mean of 0 and a standard deviation of 1, which improves the stability and efficiency of the learning process. To achieve this, we used the StandardScaler method from the scikit-learn Python library.  

Let a time series of length \( N \) be represented as  
\( V_i, \quad i = 1, 2, \dots, N \).  The equation for standardization is as follows
\begin{equation}
    V_i' = \frac{V_i - \mu}{\sigma}
\end{equation}

\noindent where \( V_i' \) represents the standardized value, \( V_i \) denotes the observed values in the dataset, \( \mu \) is the mean of \( V \), and \( \sigma \) is the standard deviation of \( V \).  
\subsection{Modeling}
\subsubsection{Structuring time series data for supervised learning}
To adapt time series data for supervised learning, it is first transformed into structured instances with predefined input and output attributes. These instances are then divided into separate training and test sets to facilitate model learning and evaluation. This step ensures the construction of appropriate datasets for training, testing, and validation.

\subsubsection{Sequential data preparation for LSTM}  
LSTM networks are designed to handle sequential data, making them suitable for time series forecasting. To effectively utilize LSTM, we need to transform the dataset into sequences of fixed-length time steps.  
To effectively train an LSTM model for time series forecasting, we structure the dataset in two phases, each designed to capture different levels of temporal dependencies.
\begin{itemize}
    \item \textit{Phase 1 (Intraday demand sequence for LSTM)}

\noindent In this phase, we represent time series demand data in structured daily groups, where each day consists of five discrete time slots. Let the demand for the day  be represented as

\[
V_d = \{ V_{d,1}, V_{d,2}, V_{d,3},V_{d,4},V_{d,5} \}
\]

where \( V_{d,t} \) denotes the demand at time slot t on day d. To construct overlapping sequences for training and prediction, we segment the data into input-output pairs using a sliding window approach over n days.
\[
X_d = [V_{d,1}, V_{d,2}, V_{d,3}, V_{d,4}, V_{d,5}, V_{d+1,1}, \dots V_{d+n-1,5}]
\]
where \(X_d\) captures a total of 5n demand values. The corresponding target sequence \(Y_d\) represents the demand values for the next day (d+n) across all five time slots
\[
Y_d = [V_{d+n,1},V_{d+n,2},V_{d+n,3},V_{d+n,4},V_{d+n,5}]
\]

This approach ensures that each input \(X_d\) includes sufficient historical context, while the target \(Y_d\) provides the full demand pattern for the subsequent day.

\item \textit{Phase 2 (Aggregated daily demand for LSTM training)}

In this phase, we aggregate the demand values from the five discrete time slots into a single daily demand value. This transformation simplifies the time series representation, making it suitable for models that predict daily demand trends rather than intra-day variations.

To achieve this, we compute the average demand for each day

\[
V_d^{avg} = \frac{1}{5} \sum_{t=1}^{5} V_{d,t}
\]

where \(V_d^{avg}\) represents the average demand for day \(d\).

After transforming the time series into daily averages, we prepare the dataset for LSTM training using a sliding window approach over six days. The input sequence for each training instance consists of the average demand over six consecutive days

\[
X_d = [V_d^{avg}, V_{d+1}^{avg}, V_{d+2}^{avg}, V_{d+3}^{avg}, V_{d+4}^{avg}, V_{d+5}^{avg}]
\]

The corresponding target value represents the predicted daily demand for the seventh day

\[
Y_d = V_{d+6}^{avg}
\]

This approach ensures that the LSTM model captures longer-term temporal dependencies by learning from daily aggregated patterns. It is particularly useful for forecasting overall demand trends rather than fluctuations within a day.
\end{itemize}
\subsubsection{Splitting data into training, testing, and validation sets}
When dealing with extensive datasets, it is beneficial to divide the data into three distinct subsets: the training set (train the model by learning demand patterns from past data), the validation set (fine-tune the model by selecting the optimal hyperparameters and preventing overfitting), and the test set (evaluation on unseen data).
\subsubsection{Hyperparameter optimization}
Selecting the most suitable model configuration is a crucial and time-intensive process. Optimizing hyperparameters enhances model performance, improves generalization, and reduces reliance on manual trial and error. Since hyperparameters are independent of the dataset, they act as control settings that influence how the model learns and adapts.

A common approach for tuning hyperparameters is grid search, which systematically explores different parameter combinations to identify the best-performing configuration. However, as the number of parameters increases, grid search can become computationally expensive. In this study, we optimize the LSTM model using the following hyperparameters

\begin{itemize}
    \item {Epochs} (\(E\)): The number of times the entire dataset is passed through the model during training, which helps in adjusting model weights and improving learning. 
    \item {LSTM units} (\(U\)): The number of neurons in the LSTM layer, which determines the model's ability to capture temporal dependencies in data. 
    \item {Batch size} (\(B\)): The number of samples processed before updating model weights, influencing training speed and stability. 
    \item {Dropout rate} (\(D\)): A regularization technique that randomly deactivates a fraction of neurons during training to prevent overfitting. 
    \item {Learning rate} (\(\alpha\)): The step size at which the model updates weights during training, affecting convergence speed and accuracy. 
    \item {Optimizer}: The optimization algorithm used to update model weights and minimize errors. We use the Adam optimizer to efficiently adjust weights and reduce the mean squared error (MSE). 
    \item {Layer configuration}: The depth of the LSTM network, affecting its ability to capture complex patterns. Our model ranges from a monolayer to multiple stacked layers.
    \item {Training time}: The total time required to train the model, which depends on the dataset size, network complexity, and computational resources.
\end{itemize}

By tuning these hyperparameters, we aim to improve the LSTM model’s ability to predict demand accurately while maintaining computational efficiency.

\subsubsection{Performance assessment}
The predicted model is evaluated using a test dataset separately for both phases. The performance metrics are computed for both Swiggy and Zomato demands
:  
\begin{itemize}
    \item[a)] {Root mean squared error (RMSE)} measures the average magnitude of prediction errors, providing an indicator of model accuracy: 
\begin{equation}
RMSE = \sqrt{\frac{1}{n} \sum_{i=1}^{n} (Y_i - \hat{Y}_i)^2}
\end{equation}
A lower RMSE value indicates better model performance.  
    \item[b)] {Mean absolute error (MAE)} measures the average absolute difference between actual and predicted values, providing a straightforward interpretation of prediction errors  
\begin{equation}
MAE = \frac{1}{n} \sum_{i=1}^{n} \left| Y_i - \hat{Y}_i \right|
\end{equation}
A lower MAE indicates better model performance, as it represents smaller deviations between predictions and actual values.
    \item[c)] {R-Squared (\(R^2\)) score} evaluates the proportion of variance in the dependent variable explained by the model
\begin{equation}
R^2 = 1 - \frac{\sum (Y_i - \hat{Y}_i)^2}{\sum (Y_i - \bar{Y})^2}
\end{equation}
where \( \bar{Y} \) is the mean of the actual values. An \( R^2 \) value closer to 1 indicates a better fit of the model to the data. 
\end{itemize}
\section{Experimental framework}
This section analyses the numerical setup and experimental process of the supply chain model.
\subsection{Numerical Setup}
The dataset used in this study simulates online food delivery demand forecasting for Zomato and Swiggy, covering the period from January 2023 to January 2025. Demand is recorded at five time intervals per day—morning, noon, evening, night, and midnight—resulting in a total of 3,675 observations across 19 variables. It includes temporal attributes such as the week, date, day, and time of order placement, allowing for an analysis of time-dependent demand variations. Order characteristics, including food category, price, and customer demand, provide insights into consumer purchasing patterns on both platforms. Delivery-related metrics, such as lead time and delivery distance, reflect logistical constraints and service efficiency. Additionally, the dataset incorporates external factors that impact demand, including supplier inventory levels, public holidays, special events categorized by importance (high, medium, or low), and weather conditions.  

For the cyclical demand pattern in Equation (1), we assume $\beta_{\text{phase}} = 0.475$, with $P = 5$ time periods per day and a peak shift of $c = 1.5$ to align with dinner hours. The baseline demand is set at $h = 0.525$ to reflect realistic daily fluctuations.

The base demand values are defined as $\alpha_s(t) = 12,000$ for Swiggy and $\alpha_z(t) = 10,000$ for Zomato, based on observed market trends. The pricing structure follows average Indian food prices, and price sensitivity parameters $\beta_s(t_i)$ and $\beta_z(t_i)$ are selected based on empirical studies \cite{andreyeva2010impact}.

For order arrivals, we assume an M/G/$\infty$ queuing model, with the arrival rate $\lambda_{\text{arrival}}$ varying across demand periods

\[
\lambda_{\text{arrival}} = 
\begin{cases}
    3 - 10 & \text{peak hours} \\
    2 - 6 & \text{standard times} \\
    1 - 3 & \text{late-night hours}
\end{cases}
\]
The service time $T_{p, i}$ is assumed to follow a log normal distribution with a mean of 30 minutes and a standard deviation of 10 minutes, reflecting realistic variation in food preparation times. These values correspond to lognormal parameters $\mu=3.35$ and $\sigma=0.32$. Delivery time depends on the restaurant-to-customer distance $\Delta(t_i)$, assumed to range between 1 km and 15 km, following a uniform distribution (in minute) to reflect real world variations.\footnote{\href{https://economictimes.indiatimes.com/industry/services/retail/what-indians-ordered-from-food-to-pet-care-here-are-the-trending-deliveries-in-2024/articleshow/117236304.cms?}{\url{https://economictimes.indiatimes.com/industry/services/retail/what-indians-ordered-from-food-to-pet-\\
care-here-are-the-trending-deliveries-in-2024/articleshow/117236304.cms?}}}
\[
T_{\text{d,i}} = 
\begin{cases}
\text{U}(15, 20) & \text{if distance} < 5 \\
\text{U}(30, 35) & \text{if } 5 \leq \text{distance} < 10 \\
\text{U}(40, 45) & \text{if distance} \geq 10 \\
\end{cases}
\]
Lead time sensitivity $\tau(t)$ is taken from prior research \cite{negi2019transportation}, while competitive factors $\gamma(t) = 0.5$ and $\delta(t) = 0.5$ reflect moderate competition between Swiggy and Zomato.
 
$\epsilon_t$ follows a normal distribution with a mean of 0 and standard deviation of 20. We use $z = 1.96$, the standard safety factor for optimal order quantity calculations. The correlation coefficient $\rho = 0.93$ represents the relationship between Swiggy and Zomato demand.

Demand fluctuations are influenced by external factors, with extreme weather increasing demand by 40–60\% (\cite{liu2022effect};\cite{yao2023weather}), holidays and special events causing a 20–50\% surge \footnote{\href{https://www.qsrmagazine.com/growth/consumer-trends/whats-next-restaurant-delivery/}{\url{https://www.qsrmagazine.com/growth/consumer-trends/whats-next-restaurant-delivery/}}}, and loyal customers generating 10–20\% more orders \footnote{\href{https://www.adjust.com/blog/hunger-for-food-delivery-apps-grows-in-2022-against-all-odds/}{\url{https://www.adjust.com/blog/hunger-for-food-delivery-apps-grows-in-2022-against-all-odds/}}} compared to others. These variations highlight the importance of accurate forecasting in optimizing inventory and supply chain management.
The detailed derivation of all model parameters and assumptions is provided in Appendix
\subsection{Experiment process}
The experiment follows a structured approach to prepare and train the model effectively. The key steps include the following:

\subsubsection{Feature selection and target variable identification}
The dependent variable for demand forecasting includes Zomato Demand \( D_Z(t_i) \) and Swiggy Demand \( D_S(t_i) \), representing food orders on each platform. Independent variables include price \( P_Z(t_i) \), \( P_S(t_i) \), lead time \( L_Z(t_i) \), \( L_S(t_i) \), and distance \( \Delta_Z(t_i) \), \( \Delta_S(t_i) \). External factors like event importance \( E(t) \) and weather conditions \( W(t) \) also influence demand fluctuations.

\subsubsection{Exploratory data analysis (EDA)}
To analyze demand patterns, we compute the daily average demand to smooth short-term fluctuations:  

\begin{equation}
D_{\text{avg}, t} = \frac{1}{5} \sum_{i=1}^{5} D_{t,i}
\end{equation}

Long-term trends are captured using the cumulative mean:  

\begin{equation}
CM_t = \frac{1}{t} \sum_{j=1}^{t} D_{\text{avg}, j}
\end{equation}

Demand stability is assessed via the 7-day rolling variance, considering deviations from the rolling mean:

\begin{equation}
RV_t = \frac{1}{7} \sum_{j=t-6}^{t} (D_{\text{avg}, j} - \bar{D}_{\text{avg}, t})^2
\end{equation}

Additionally, we analyze the demand distribution using a histogram to identify skewness and outliers, and a Q-Q plot to compare the demand distribution against a normal distribution.

\subsubsection{Train-validation-test split}  

For the train-validation-test split, both Swiggy and Zomato demand data are simulated from January 1, 2023, to January 1, 2025. To ensure a proper distribution, we allocate 70\% of the data (January 2023 – April 2024) for training, 15\% (May 2024 – September 2024) for validation, and the remaining 15\% (October 2024 – January 2025) for testing.  

\subsubsection{Hyperparameter optimization via GridSearch}
Table \ref{tab:table2} presents the selected values for each hyperparameter, which will be used for configuring and training the model in both the phases. These values have been determined based on prior experimentation and literature review to ensure optimal performance.

\begin{table}[h]
    \centering
    \begin{tabular}{|l|c|}
        \hline
        Hyperparameter & Value Range \\
        \hline
        Epochs (\(E\)) & 50 – 200 \\
        LSTM Units (\(U\)) & 32 – 128 \\
        Batch Size (\(B\)) & 16 – 64 \\
        Dropout Rate (\(D\)) & 0.1 – 0.5 \\
        Learning Rate (\(\alpha\)) & 0.001 – 0.01 \\
        Layers  & 1-3 \\
        \hline
    \end{tabular}
    \caption{Hyperparameter values for GridSearch}
    \label{tab:table2}
\end{table}

\subsubsection{Bullwhip effect measurements}
The calculation is performed in three stages: First, the bullwhip effect is computed using the training dataset, establishing a baseline measurement of demand fluctuations in historical data. Next, the same calculation is performed on the test dataset for Zomato and Swiggy demand, along with restaurant inventory for both phases. Lastly, calculate bullwhip effects on the predicted dataset after forecasting demand, with the inventory adjusted for both individual platforms and overall for both phases.

\section{Results and discussion}
This section presents the results of the demand forecasting model and its impact on supply chain stability. The analysis follows a structured approach:

\subsection{EDA results}

\begin{figure}[H]
    \centering
    \begin{subfigure}[b]{0.7\textwidth}
        \centering
        \includegraphics[width=\textwidth]{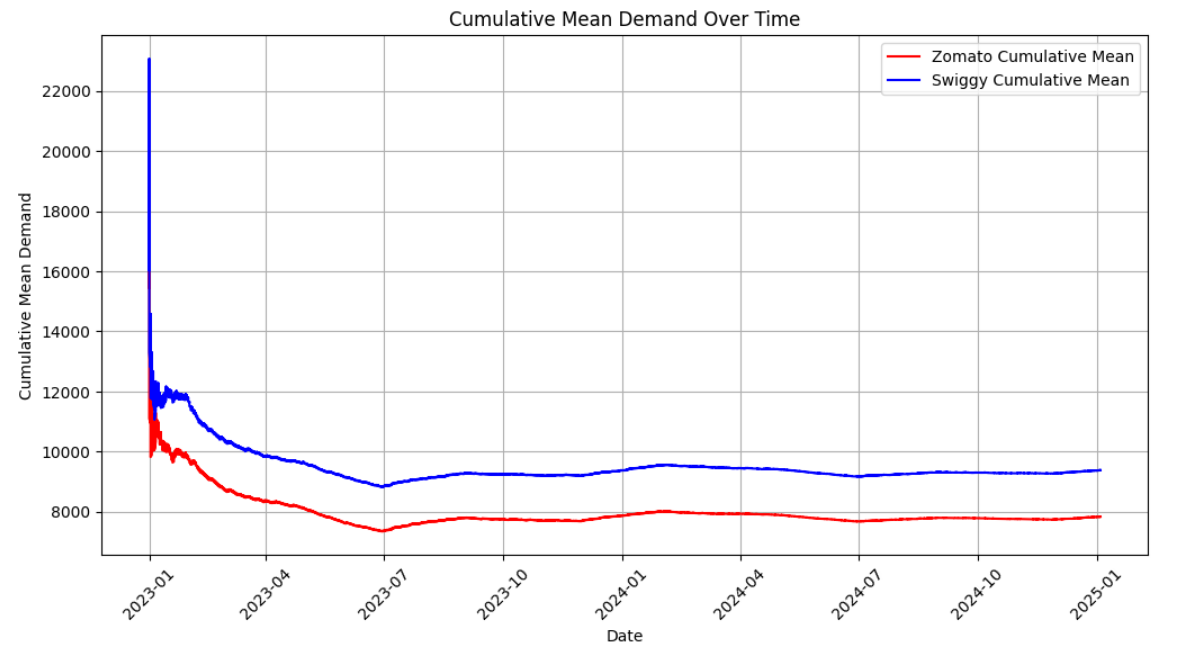}
        \caption{Cumulative mean of demand over time}
        \label{fig:cumulative_mean}
    \end{subfigure}
\vskip 1cm
    \begin{subfigure}[b]{0.7\textwidth}
        \centering
        \includegraphics[width=\textwidth]{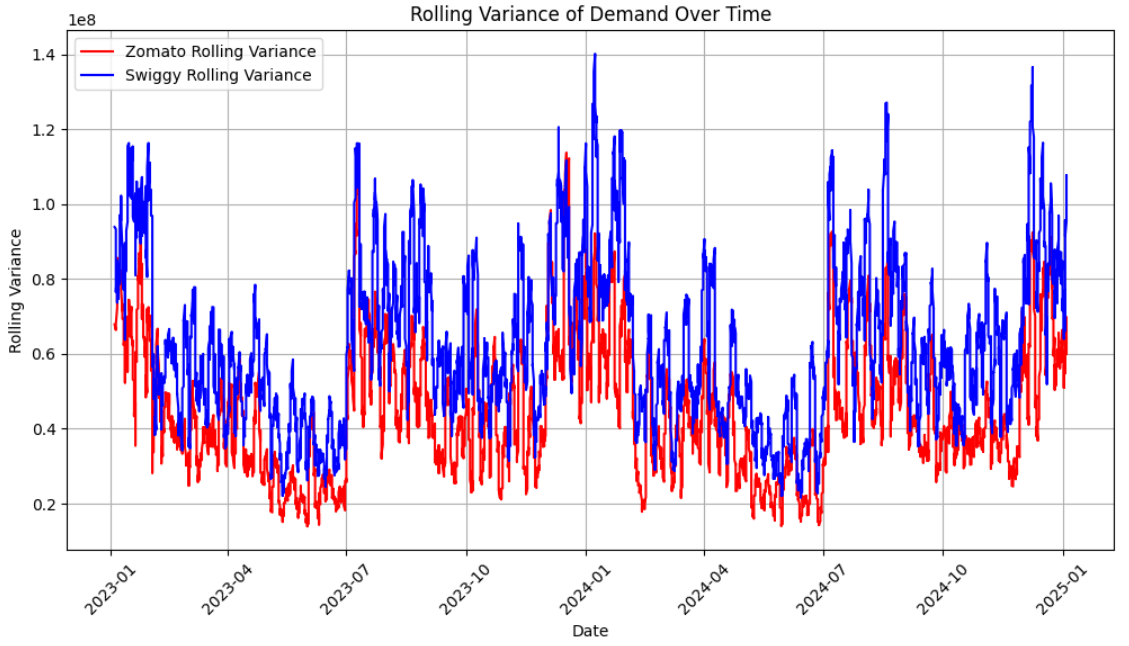}
        \caption{Rolling variance of demand over time}
        \label{fig:rolling_var}
    \end{subfigure}
    \caption{Cumulative mean and rolling variance of demand for Zomato and Swiggy}
    \label{fig:eda_results}
\end{figure}

Figure \ref{fig:eda_results} presents the cumulative mean and rolling variance of demand for Zomato and Swiggy over time. These visualizations help in understanding long-term trends and demand fluctuations.  

\begin{figure}[H]
    \centering
    \includegraphics[width=0.9\linewidth]{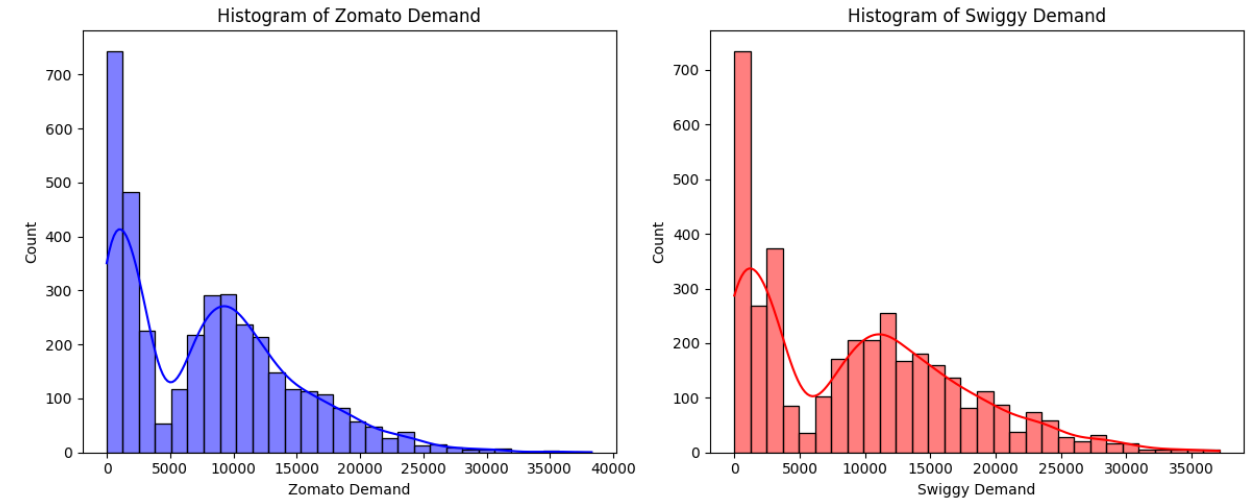}
    \caption{Histogram of demand}
    \label{fig:normality}
\end{figure}

\begin{figure}[ht]
    \centering
    \includegraphics[width=0.9\linewidth]{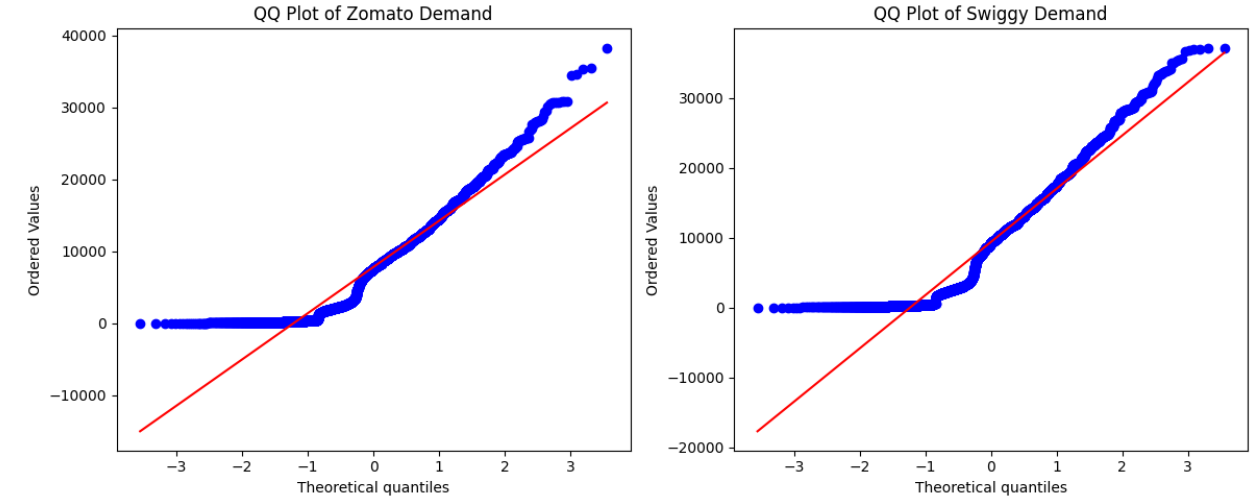}
    \caption{QQ plot of demand}
    \label{fig:qqplot}
\end{figure}

\noindent The cumulative mean of demand, shown in Figure \ref{fig:cumulative_mean}, highlights the overall trend for both platforms. Initially, Swiggy exhibits a higher average demand than Zomato, but both platforms stabilize over time, with Zomato settling at a lower demand level. The rolling variance, presented in Figure \ref{fig:rolling_var}, reveals fluctuations in daily demand. Swiggy exhibits consistently higher volatility compared to Zomato, indicating greater variability in its order patterns. These findings provide key insights into demand stability, helping to refine forecasting models. Additionally, to assess the distribution of demand, we plot the Histogram \& QQ Plot given in Figures \ref{fig:normality} and \ref{fig:qqplot}.

Both Zomato and Swiggy demand exhibit a bimodal distribution in Figure, indicating two distinct demand clusters. The positive skewness suggests occasional periods of significantly higher demand, requiring strategic resource planning.
The QQ plots provide visual evidence that the demand for both Zomato and Swiggy deviates from a normal distribution, particularly in the tails, suggesting the presence of more frequent extreme values than a normal distribution would predict.

\subsection{Grid search \& hyperparameter tuning}
For both phase 1  and phase 2, we perform hyperparameter tuning using Grid Search on the validation dataset, evaluating various hyperparameter combinations mentioned in Table \ref{tab:table2} to identify the optimal configuration that minimizes the validation loss.  
 \begin{figure}[ht]
     \centering
     \includegraphics[width=0.8\linewidth]{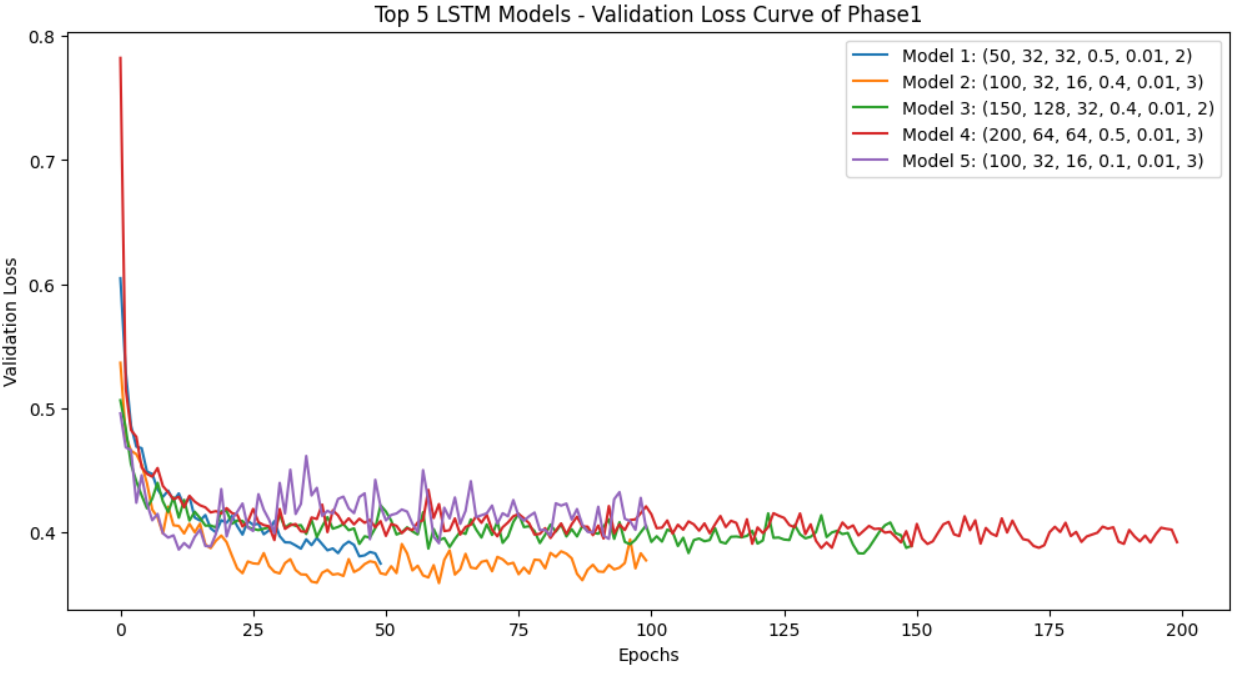}
     \caption{Validation loss curve for phase 1 (intraday demand prediction)}
     \label{fig:valp1}
 \end{figure}
\begin{figure}[ht]
    \centering
    \includegraphics[width=0.8\linewidth]{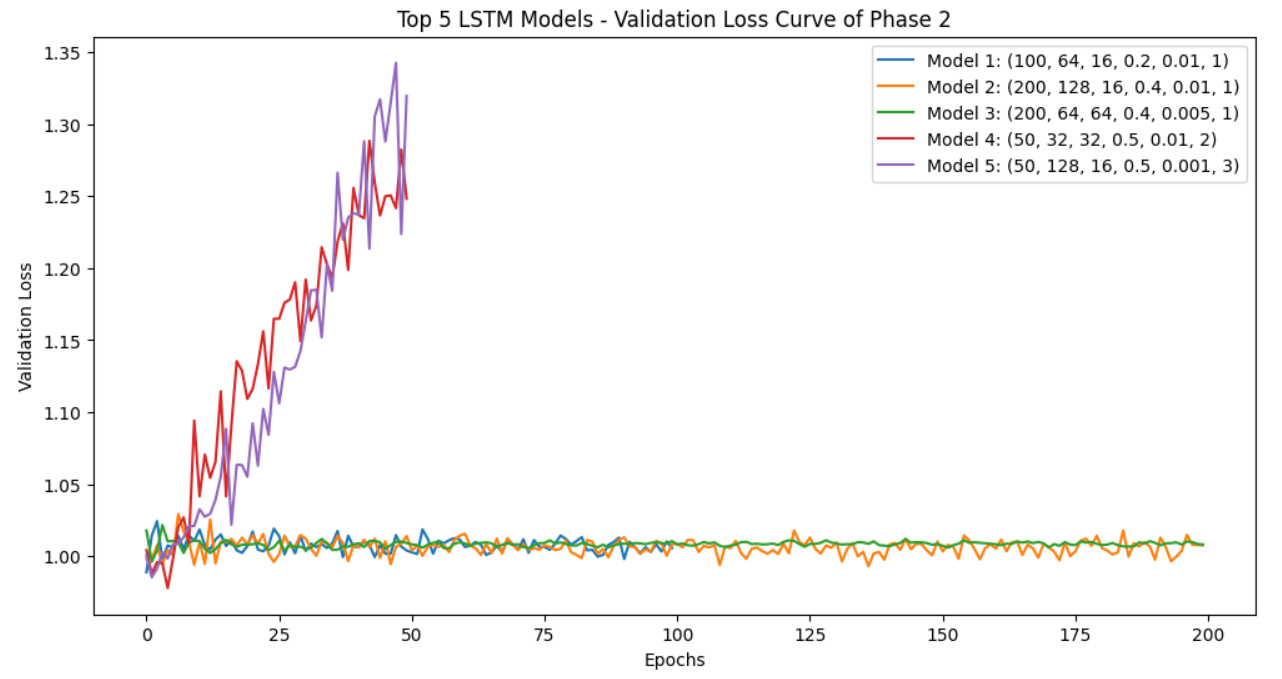}
    \caption{Validation loss curve for phase 2 (daily aggregated demand prediction)}
    \label{fig:valp2}
\end{figure} 
Figure \ref{fig:valp1} and Figure \ref{fig:valp2} suggest {model $2$} is the best. It shows the lowest and most stable validation loss over the epochs, indicating it generalizes the best to unseen data for both phases.
Using the best hyperparameter combination, we predict future demand for both Swiggy and Zomato across both phase $1$ and phase $2$. 
\newpage
\subsection{Performance analysis}
\begin{figure}[H]
    \centering
    \includegraphics[width=0.9\linewidth]{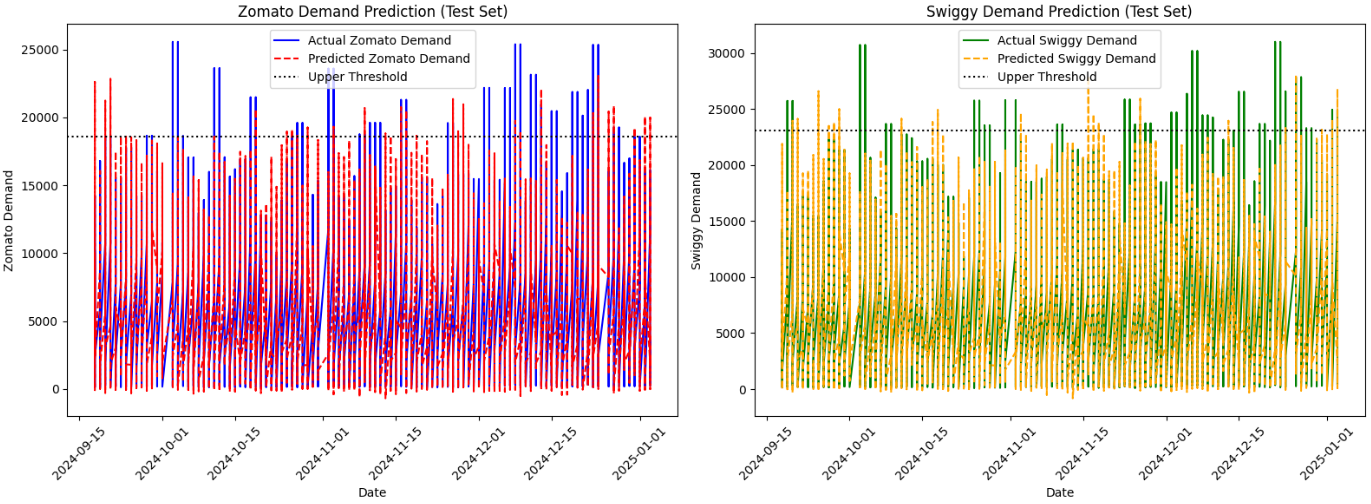}
    \caption{Actual vs. predicted demand for phase 1 (intraday demand prediction)}
    \label{fig:p1}
\end{figure}

\begin{figure}[H]
    \centering
    \includegraphics[width=0.99\textwidth, height=1\textheight, keepaspectratio]{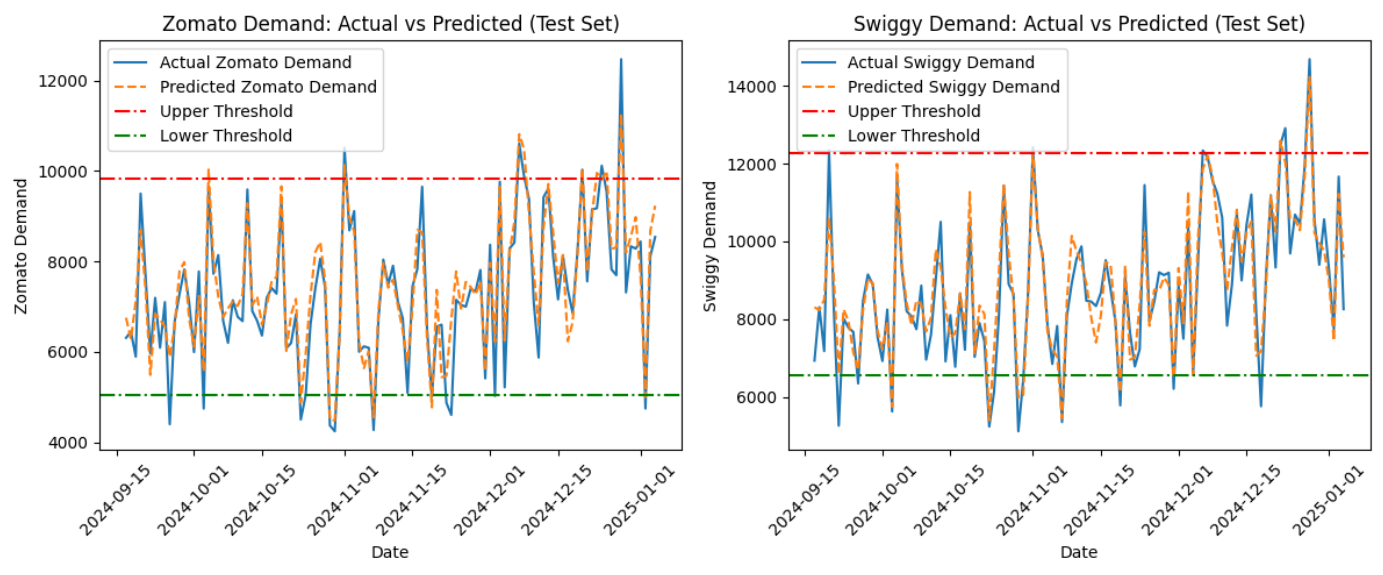}
    \caption{Actual vs. predicted demand for phase 2 (daily aggregated demand prediction)}
    \label{fig:p2}
\end{figure}

Figure \ref{fig:p1}, \ref{fig:p2} present the actual and predicted demand plots for the test dataset, demonstrating that the LSTM model effectively captures complex nonlinear patterns in each case.  

After predicting future demand for both Swiggy and Zomato, we analyze the impact on restaurant inventory. The inventory levels are adjusted based on the forecasted demand to ensure optimal stock availability while minimizing wastage.  
\begin{figure}[H]
    \centering
    \includegraphics[width=0.8\linewidth]{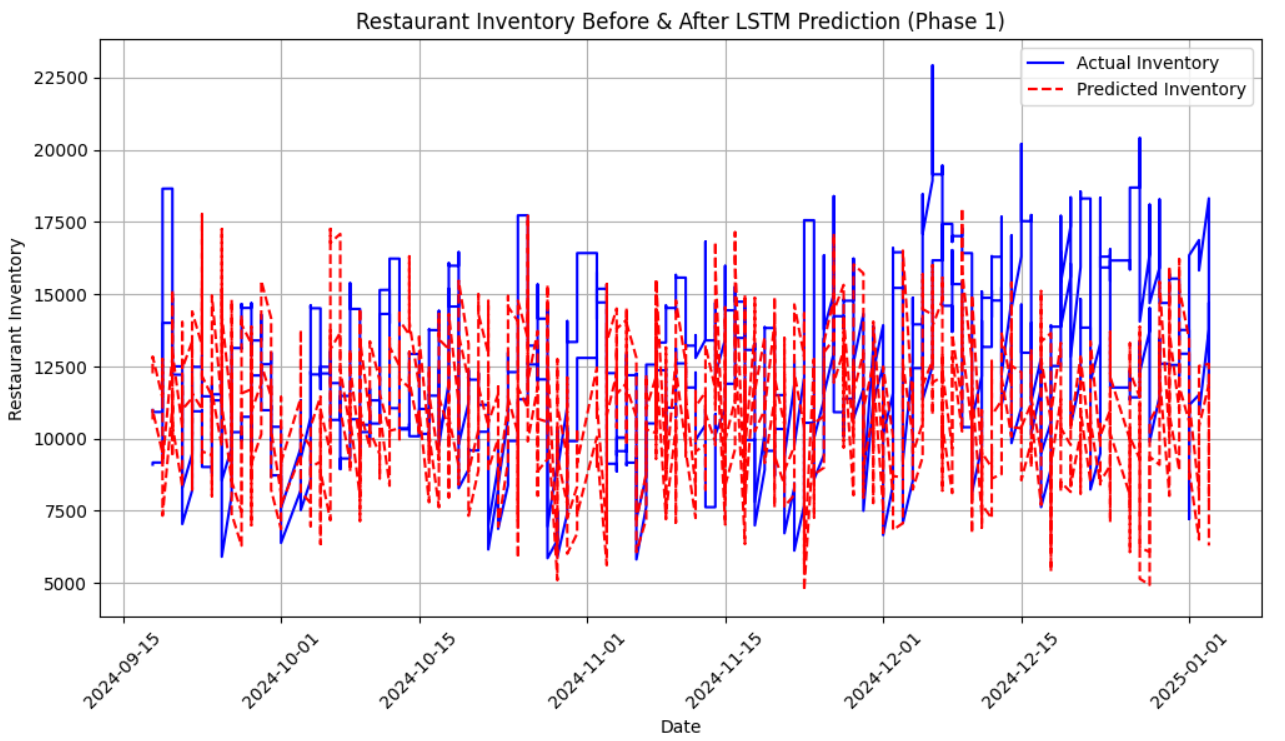}
    \caption{Restaurant inventory levels after predictive demand for phase 1}
    \label{fig:inventory_phase1}
\end{figure}
\begin{figure}[H]
    \centering
    \includegraphics[width=0.8\linewidth]{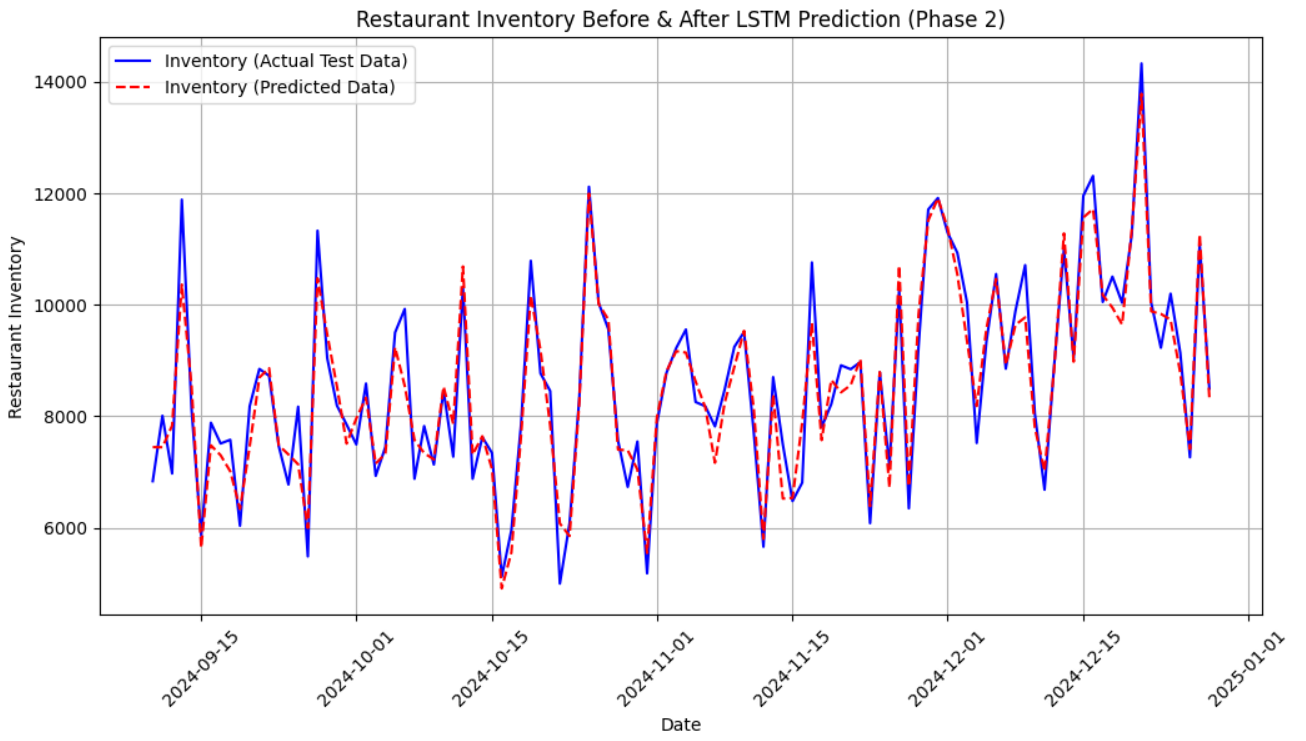}
    \caption{Restaurant inventory levels after predictive demand for phase 2}
    \label{fig:inventory_phase2}
\end{figure}
Figure \ref{fig:inventory_phase1} and Figure \ref{fig:inventory_phase2} illustrate the restaurant inventory trends after incorporating predictive demand insights for both phases, respectively. These plots show how LSTM smooths out demand uncertainty, leading to a more stable inventory that aligns with the predicted demand. Table \ref{tab:evaluation_metrics} presents the evaluation metrics for both phases, which measure the model's accuracy in predicting demand for Swiggy and Zomato.
\begin{table}[ht]
    \centering
    \renewcommand{\arraystretch}{1.3} 
    \setlength{\tabcolsep}{8pt} 
    \begin{tabular}{|c|c|c|c|c|c|}
        \hline
        Phase& Platform & RMSE & MAE & R\textsuperscript{2} & Training Time(minute)\\
        \hline
        \multirow{2}{*}{Phase 1} & Zomato & 5531.67 & 4130.31 & 0.69 & 12 \\
                                          & Swiggy & 6768.82 & 4992.76 & 0.71 & 12 \\
        \hline
        \multirow{2}{*}{Phase 2} & Zomato & 536.25 & 414.51 & 0.88 & 8\\
                                          & Swiggy & 579.40 & 468.78 & 0.90 & 8\\
        \hline
    \end{tabular}
    \caption{Evaluation metrics for phase 1 and phase 2}
    \label{tab:evaluation_metrics}
\end{table}

\subsection{Measurement of bullwhip effect}
We calculate the bullwhip effect for a restaurant through Zomato and Swiggy across different data segments, including training data, test data, and predicted demand, using LSTM with inventory smoothing. As shown in Table \ref{tab:bullwhip_effect}, a high reduction in the bullwhip effect suggests improved demand stability. This analysis is conducted separately for both phases to evaluate how short-term and long-term demand variations impact inventory management.

\begin{table}[ht]
    \centering
    \renewcommand{\arraystretch}{1.9} 
    \setlength{\tabcolsep}{5pt} 
    \label{tab:bullwhip_effect}
    \begin{tabular}{|c|ccc|ccc|ccc|}
        \hline
        \multirow{2}{*}{Phase} & \multicolumn{3}{c|}{Training} & \multicolumn{3}{c|}{Testing} & \multicolumn{3}{c|}{Predicted} \\
        \cline{2-10}
        & Zomato & Swiggy & Overall & Zomato & Swiggy & Overall & Zomato & Swiggy & Overall \\
        \hline
        Phase 1 & 2.88 & 2.35 & 2.61 & 2.06 & 1.96 & 2.01 & 1.02 & 0.91 & 0.96 \\
        \hline
        Phase 2 & 2.32 & 2.07 & 2.19 & 1.97 & 1.76 & 1.86 & 0.72 & 0.89 & 0.80 \\
        \hline
    \end{tabular}
     \caption{Bullwhip effect analysis across phase 1 and phase 2}
     \label{tab:bullwhip_effect}
\end{table}

\section{Conclusion \& managerial insights}
The primary objective of this study was to develop an LSTM-based demand forecasting model to improve both intra-day and aggregated daily demand predictions for restaurants, utilizing real-world sales data from Swiggy and Zomato. By addressing demand fluctuations, the model aimed to enhance inventory planning and reduce the bullwhip effect, a common challenge in restaurant supply chains. The results validate the effectiveness of neural networks in mitigating these inefficiencies, aligning with findings from \citep{rezaeefard2024present}, which demonstrated the superiority of LSTM over traditional forecasting models. To achieve optimal performance, the model was fine-tuned using the grid search method, ensuring it captured the complex patterns and seasonality present in the time-series data. A lower bullwhip effect indicates a more stable and predictable supply chain, minimizing inefficiencies such as overstocking or shortages. The results show that in Phase 1, the bullwhip effect decreased from 2.61 to 0.96, with $R^2$ values of 0.69 for Zomato and 0.71 for Swiggy. While this demonstrates a significant reduction in demand variability, Phase 2 performed even better, with the bullwhip effect further declining from 2.19 to 0.80. The higher $R^2$ values of 0.88 for Zomato and 0.90 for Swiggy in Phase 2 indicate improved prediction accuracy, meaning the model was better at explaining demand variations. In conclusion, the results demonstrate that LSTM-based demand forecasting, combined with inventory smoothing, is an effective approach to minimizing the bullwhip effect in restaurant supply chains. The proposed model achieved strong predictive accuracy, and its integration into supply chain management can significantly enhance efficiency and improve decision-making for restaurants relying on customer orders placed through Swiggy and Zomato. 

The findings of this research have significant practical implications for restaurant operators, supply chain managers, and online food delivery platforms. By implementing the proposed LSTM-based demand forecasting model with inventory smoothing, these stakeholders can enable them to make more informed decisions. For restaurant operators, accurate intraday forecasting ensures optimal ingredient availability during peak hours while minimizing waste during low-demand periods. At the aggregate daily level, it enables better procurement planning, reducing last-minute orders and supply chain disruptions. For supply chain managers, improved intraday demand predictions allow for better synchronization of supplier deliveries with restaurant needs, reducing storage costs and ensuring a steady supply of perishable goods. On a daily scale, stable demand forecasts lead to more efficient production planning and cost-effective logistics, minimizing uncertainty in order quantities. For online food delivery platforms, precise intraday demand forecasting helps allocate delivery resources efficiently, reducing wait times and improving customer satisfaction. At the aggregate level, demand insights optimize dynamic pricing, targeted promotions, and restaurant recommendations, driving higher order volumes and platform engagement. By leveraging LSTM-based forecasting for both intraday and daily demand, these stakeholders can enhance efficiency, reduce costs, and improve overall supply chain stability, ensuring long-term sustainability in the competitive food delivery industry.

While this study demonstrates the effectiveness of LSTM-based forecasting in reducing the bullwhip effect and improving demand prediction for restaurants, certain limitations remain. One key gap is that due to the unavailability of industry used data, the study relies on simulated data rather than real-world transactional data from restaurants and food delivery platforms. Although the simulation effectively captures realistic demand fluctuations, some assumptions are made intuitively due to the lack of relevant literature. Additionally, while the model successfully predicts demand and mitigates the bullwhip effect, it does not optimize inventory replenishment in real time, which is crucial for minimizing costs and ensuring supply chain efficiency.

To address these gaps, future research should focus on implementing the model using real-world datasets from restaurants and food delivery platforms to validate its practical effectiveness. Furthermore, integrating reinforcement learning or optimization algorithms could enhance real-time inventory management by dynamically adjusting order quantities based on demand predictions.

\vskip .7cm
\section*{\centering \underline{Appendix} } 

\textbf{A1. Cyclical Demand Patterns}  
\vskip .2cm
\noindent Food delivery demand follows a cyclical pattern, peaking during dinner hours (7 PM – 9 PM) and reaching its lowest point in the early morning (3 AM – 5 AM). On New Year's Eve, Swiggy and Zomato processed 9,500 and 4,254 orders per minute, respectively.\footnote{Swiggy receives 9,500 orders per minute on New Year's Eve - \href{https://www.business-standard.com}{\url{https://www.business-standard.com}}},\footnote{Zomato records 4,254 orders per minute on New Year's Eve\href{https://m.economictimes.com}{https://m.economictimes.com}} On a normal day, peak demand is four times lower, with Swiggy handling 2,375 orders/min and Zomato 1,064 orders/min. Off-peak demand is approximately 5\% of peak demand (Swiggy: 119 orders/min, Zomato: 53 orders/min).  

Over a 120-minute peak window, Swiggy processes 285,000 orders, while Zomato processes 127,680 orders. During off-peak periods, order volumes drop to 14,280 and 6,360 orders, respectively. The amplitude of demand fluctuations is calculated as:  

\begin{equation}
B_{\text{Swiggy}} = \frac{285,000 - 14,280}{2 \times 285,000} = 0.475, \quad B_{\text{Zomato}} = \frac{127,680 - 6,360}{2 \times 127,680} = 0.475
\end{equation}

The corresponding baseline demand levels are:  

\begin{equation}
h_{\text{Swiggy}} = \frac{285,000 + 14,280}{2 \times 285,000} = 0.525, \quad h_{\text{Zomato}} = \frac{127,680 + 6,360}{2 \times 127,680} = 0.525
\end{equation}

Dividing the day into five periods (\( P = 5 \)) and aligning the peak at dinner time (\( c = 1.5 \)), the sinusoidal demand model is given by  

\begin{equation}
\gamma_t = 0.475 \sin\left(\frac{\pi}{5} t + 1.5\right) + 0.525
\end{equation}

This formulation ensures that peak demand occurs at dinner time and declines during the early morning hours.  
\vskip .3cm
\noindent \textbf{A2. External demand multipliers}  
\vskip .2cm
\noindent The numerical values for demand multipliers are estimated based on empirical research, market trends, and food delivery platform reports. These multipliers capture variations due to time of day, weather conditions, special events, and customer behavior.  
\vskip .4cm
 \textit{Time of day multiplier:}  
\begin{equation}
T(t) = 
\begin{cases}
    1.2 - 1.5 & \text{for peak hours} \\
    1.0 - 1.2 & \text{for off-peak hours} \\
    0.8 - 1.0 & \text{for late-night hours}
\end{cases}
\end{equation}

 \textit{Weather influence multiplier:}  
\begin{equation}
W(t) = 
\begin{cases}
    1.4 - 1.6 & \text{for extreme weather conditions} \\
    1.1 - 1.3 & \text{for mild weather} \\
    1.0 & \text{for clear weather}
\end{cases}
\end{equation}
\vskip .2cm
 \textit{Holiday and special event multiplier:}  
\begin{equation}
E(t) = 
\begin{cases}
    1.2 - 1.5 & \text{for major holidays and special occasions} \\
    1.0 & \text{for regular days}
\end{cases}
\end{equation}
\vskip .2cm
 \textit{Customer segmentation multiplier:}  
\begin{equation}
\beta_{\text{customer}} = 
\begin{cases}
    0.7 - 0.9 & \text{for mismatched preferences} \\
    1.1 - 1.2 & \text{for loyal customers} \\
    1.0 & \text{for general customers}
\end{cases}
\end{equation}

\vskip .2cm
\noindent \textbf{A3. Service time food preparation modeling}  
\vskip .2cm
\noindent The service time \( T_{p,i} \) follows a lognormal distribution with a mean of 30 minutes and a standard deviation of 10 minutes. The mean and standard deviation of a lognormal distribution are given by the formulas
\[
m = e^{\mu + \frac{\sigma^2}{2}} = 30
\]
\[
s = \sqrt{(e^{\sigma^2} - 1) e^{2\mu + \sigma^2}} = 10
\]
To determine the lognormal parameters \( \mu \) and \( \sigma \), we use the standard transformations. The location parameter \( \mu \) is given by:

\[
\mu = \ln \left( \frac{m^2}{\sqrt{s^2 + m^2}} \right)
\]

Substituting \( m = 30 \) and \( s = 10 \), we obtain:
\[
\mu = \ln \left( \frac{30^2}{\sqrt{10^2 + 30^2}} \right) = \ln(28.47) \approx 3.35
\]
Similarly, the scale parameter is computed as
\[
\sigma^2 = \ln \left( 1 + \frac{s^2}{m^2} \right)
\]

which evaluates to:

\[
\sigma^2 = \ln(1.111) \approx 0.105
\]

Taking the square root:

\[
\sigma = \sqrt{0.105} \approx 0.32
\]

Thus, the service time is modeled as:

\[
T_{p,i} \sim \text{Lognormal}(3.35, 0.32)
\]

\vskip .3cm
\noindent \textbf{A4. Baseline Demand and Market Correlation}  
\vskip .2cm
\noindent We simulate daily demand over two years at five time intervals per day. The baseline demand values for Swiggy and Zomato are determined using their relative market shares. Reports indicate that Swiggy’s order volume was 1.2 times higher than Zomato's during this period \footnote{\href{https://www.vumonic.com/blog/showdown-of-food-delivery-giants-the-market-share-aov-consumer-behavior-of-swiggy-and-zomato-from-april-june-2023}{Swiggy-Zomato Market Share Analysis}} while Zomato processed over 647 million orders in the financial year.\footnote{\href{https://www.statista.com/statistics/1110238/zomato-number-of-orders}{Zomato Order Statistics (2019-2023)}}  

\indent Based on these observations, the baseline demand values are defined as 

\begin{equation}
\alpha_s(t) = 12,000, \quad \alpha_z(t) = 10,000
\end{equation}

\noindent The correlation coefficient \( \rho \) between Swiggy and Zomato demand is calculated as:

\begin{equation}
\rho = \frac{\sum (X_i - \bar{X})(Y_i - \bar{Y})}{\sqrt{\sum (X_i - \bar{X})^2} \sqrt{\sum (Y_i - \bar{Y})^2}}
\end{equation}

\noindent For this dataset, the computed correlation value is \( \rho = 0.93 \), indicating a strong positive relationship between demand trends for both platforms.

\vskip .5cm
\bibliographystyle{harvard}
\bibliography{reference}

@incollection{forrester2012industrial,
  title={Industrial dynamics: a major breakthrough for decision makers},
  author={Forrester, Jay W},
  booktitle={The Roots of Logistics},
  pages={141--172},
  year={2012},
  publisher={Springer}
}

@article{forrester1997industrial,
  title={Industrial dynamics},
  author={Forrester, Jay Wright},
  journal={Journal of the Operational Research Society},
  volume={48},
  number={10},
  pages={1037--1041},
  year={1997},
  publisher={Taylor \& Francis}
}

@article{sterman1989modeling,
  title={Modeling managerial behavior: Misperceptions of feedback in a dynamic decision making experiment},
  author={Sterman, John D},
  journal={Management science},
  volume={35},
  number={3},
  pages={321--339},
  year={1989},
  publisher={INFORMS}
}

@article{lee2000value,
  title={The value of information sharing in a two-level supply chain},
  author={Lee, Hau L and So, Kut C and Tang, Christopher S},
  journal={Management science},
  volume={46},
  number={5},
  pages={626--643},
  year={2000},
  publisher={INFORMS}
}

@article{lee1997information,
  title={Information distortion in a supply chain: The bullwhip effect},
  author={Lee, Hau L and Padmanabhan, Venkata and Whang, Seungjin},
  journal={Management science},
  volume={43},
  number={4},
  pages={546--558},
  year={1997},
  publisher={Informs}
}

@article{lee1997bullwhip,
  title={The bullwhip effect in supply chains},
  author={Lee, Hau L and Padmanabhan, V and Whang, Seungjin},
  year={1997}
}

@article{sodhi2012researchers,
  title={Researchers' perspectives on supply chain risk management},
  author={Sodhi, ManMohan S and Son, Byung-Gak and Tang, Christopher S},
  journal={Production and operations management},
  volume={21},
  number={1},
  pages={1--13},
  year={2012},
  publisher={SAGE Publications Sage CA: Los Angeles, CA}
}

@article{ramanathan2014performance,
  title={Performance of supply chain collaboration--A simulation study},
  author={Ramanathan, Usha},
  journal={Expert Systems with Applications},
  volume={41},
  number={1},
  pages={210--220},
  year={2014},
  publisher={Elsevier}
}

@article{yao2008inventory,
  title={The inventory value of information sharing, continuous replenishment, and vendor-managed inventory},
  author={Yao, Yuliang and Dresner, Martin},
  journal={Transportation Research Part E: Logistics and Transportation Review},
  volume={44},
  number={3},
  pages={361--378},
  year={2008},
  publisher={Elsevier}
}

@article{bai2020supply,
  title={A supply chain transparency and sustainability technology appraisal model for blockchain technology},
  author={Bai, Chunguang and Sarkis, Joseph},
  journal={International journal of production research},
  volume={58},
  number={7},
  pages={2142--2162},
  year={2020},
  publisher={Taylor \& Francis}
}

@article{costantino2015spc,
  title={SPC forecasting system to mitigate the bullwhip effect and inventory variance in supply chains},
  author={Costantino, Francesco and Di Gravio, Giulio and Shaban, Ahmed and Tronci, Massimo},
  journal={Expert Systems with Applications},
  volume={42},
  number={3},
  pages={1773--1787},
  year={2015},
  publisher={Elsevier}
}

@article{chen2012bullwhip,
  title={Bullwhip effect measurement and its implications},
  author={Chen, Li and Lee, Hau L},
  journal={Operations research},
  volume={60},
  number={4},
  pages={771--784},
  year={2012},
  publisher={INFORMS}
}

@article{yang2021behavioural,
  title={The behavioural causes of bullwhip effect in supply chains: A systematic literature review},
  author={Yang, Y and Lin, J and Liu, G and Zhou, L},
  journal={International Journal of Production Economics},
  volume={236},
  pages={108120},
  year={2021},
  publisher={Elsevier}
}

@article{assad2023comparing,
  title={Comparing short-term univariate and multivariate time-series forecasting models in infectious disease outbreak},
  author={Assad, Daniel Bouzon Nagem and Cara, Javier and Ortega-Mier, Miguel},
  journal={Bulletin of Mathematical Biology},
  volume={85},
  number={1},
  pages={9},
  year={2023},
  publisher={Springer}
}

@article{babai2022demand,
  title={Demand forecasting in supply chains: a review of aggregation and hierarchical approaches},
  author={Babai, M Zied and Boylan, John E and Rostami-Tabar, Bahman},
  journal={International Journal of Production Research},
  volume={60},
  number={1},
  pages={324--348},
  year={2022},
  publisher={Taylor \& Francis}
}

@article{sariciouglu2024analyzing,
  title={Analyzing One-Step and Multi-Step Forecasting to Mitigate the Bullwhip Effect and Improve Supply Chain Performance},
  author={Sar{\i}c{\i}o{\u{g}}lu, Alper and Genevois, M{\"u}jde Erol and Cedolin, Michele},
  journal={IEEE Access},
  year={2024},
  publisher={IEEE}
}

@article{rezki2024deep,
  title={Deep learning hybrid models for effective supply chain risk management: mitigating uncertainty while enhancing demand prediction},
  author={Rezki, Nisrine and Mansouri, Mohamed},
  journal={Acta Logistica},
  volume={11},
  number={4},
  pages={589--604},
  year={2024},
  publisher={4S go, sro}
}

@inproceedings{kiuchi2024recurrent,
  title={Recurrent Neural Network Based Reinforcement Learning for Inventory Control with Agent-based Supply Chain Simulator},
  author={Kiuchi, Atsuki},
  booktitle={2024 IEEE 20th International Conference on Automation Science and Engineering (CASE)},
  pages={1903--1909},
  year={2024},
  organization={IEEE}
}

@article{borucka2023seasonal,
  title={Seasonal methods of demand forecasting in the supply chain as support for the company’s sustainable growth},
  author={Borucka, Anna},
  journal={Sustainability},
  volume={15},
  number={9},
  pages={7399},
  year={2023},
  publisher={MDPI}
}

@article{zhu2021demand,
  title={Demand forecasting with supply-chain information and machine learning: Evidence in the pharmaceutical industry},
  author={Zhu, Xiaodan and Ninh, Anh and Zhao, Hui and Liu, Zhenming},
  journal={Production and Operations Management},
  volume={30},
  number={9},
  pages={3231--3252},
  year={2021},
  publisher={SAGE Publications Sage CA: Los Angeles, CA}
}

@misc{peng2014manage,
  title={How to manage the bullwhip effect in the supply chain: A case study on Chinese Haier Group},
  author={Peng, Ronghe and Xiao, Yi},
  year={2014}
}

@article{chen2000impact,
  title={The impact of exponential smoothing forecasts on the bullwhip effect},
  author={Chen, Frank and Ryan, Jennifer K and Simchi-Levi, David},
  journal={Naval Research Logistics (NRL)},
  volume={47},
  number={4},
  pages={269--286},
  year={2000},
  publisher={Wiley Online Library}
}

@article{ravichandran2008managing,
  title={Managing bullwhip effect: two case studies},
  author={Ravichandran, N},
  journal={Journal of Advances in Management Research},
  volume={5},
  number={2},
  pages={77--87},
  year={2008},
  publisher={Emerald Group Publishing Limited}
}

@inproceedings{diaz2022simulated,
  title={Simulated-Based Analysis of Recovery Actions Under Vendor-Managed Inventory Amid Black Swan Disruptions in the Semiconductor Industry: A Case Study from Infineon Technologies AG},
  author={Diaz, Manuel Fernando Lopera and Ehm, Hans and Ismail, Abdelgafar},
  booktitle={2022 Winter Simulation Conference (WSC)},
  pages={3513--3524},
  year={2022},
  organization={IEEE}
}

@article{dejonckheere2003measuring,
  title={Measuring and avoiding the bullwhip effect: A control theoretic approach},
  author={Dejonckheere, Jeroen and Disney, Stephen M and Lambrecht, Marc R and Towill, Denis R},
  journal={European journal of operational research},
  volume={147},
  number={3},
  pages={567--590},
  year={2003},
  publisher={Elsevier}
}

@article{li2024mitigating,
  title={Mitigating transient bullwhip effects under imperfect demand forecasts},
  author={Li, Sarah HQ and D{\"o}rfler, Florian},
  journal={arXiv preprint arXiv:2404.01090},
  year={2024}
}

@article{chopra2001strategy,
  title={Strategy, planning, and operation},
  author={Chopra, Sunil and Meindl, Peter},
  journal={Supply Chain Management},
  volume={15},
  number={5},
  pages={71--85},
  year={2001},
  publisher={Springer Berlin}
}

@article{hwang2020effects,
  title={The effects of expected benefits on image, desire, and behavioral intentions in the field of drone food delivery services after the outbreak of COVID-19},
  author={Hwang, Jinsoo and Kim, Hyunjoon},
  journal={Sustainability},
  volume={13},
  number={1},
  pages={117},
  year={2020},
  publisher={MDPI}
}

@article{suali2024role,
  title={The role of digital platforms in e-commerce food supply chain resilience under exogenous disruptions},
  author={Suali, Arunpreet Singh and Srai, Jagjit Singh and Tsolakis, Naoum},
  journal={Supply Chain Management: An International Journal},
  volume={29},
  number={3},
  pages={573--601},
  year={2024},
  publisher={Emerald Publishing Limited}
}

@article{aamer2020data,
  title={Data analytics in the supply chain management: Review of machine learning applications in demand forecasting},
  author={Aamer, Ammar and Eka Yani, LuhPutu and Alan Priyatna, IMade},
  journal={Operations and Supply Chain Management: An International Journal},
  volume={14},
  number={1},
  pages={1--13},
  year={2020}
}

@article{chu2023data,
  title={Data-driven optimization for last-mile delivery},
  author={Chu, Hongrui and Zhang, Wensi and Bai, Pengfei and Chen, Yahong},
  journal={Complex \& Intelligent Systems},
  volume={9},
  number={3},
  pages={2271--2284},
  year={2023},
  publisher={Springer}
}

@article{osman2023perishable,
  title={Perishable food supply chain management: Challenges and the way forward},
  author={Osman, Siddiq Abekah and Xu, Chaoyi and Akuful, Michael and Paul, Erick Robert},
  journal={Open Journal of Social Sciences},
  volume={11},
  number={7},
  pages={349--364},
  year={2023},
  publisher={Scientific Research Publishing}
}

@article{maheshwari2021internet,
  title={Internet of things for perishable inventory management systems: an application and managerial insights for micro, small and medium enterprises},
  author={Maheshwari, Pratik and Kamble, Sachin and Pundir, Ashok and Belhadi, Amine and Ndubisi, Nelson Oly and Tiwari, Sunil},
  journal={Annals of Operations Research},
  year={2021},
  publisher={SPRINGER VAN GODEWIJCKSTRAAT 30, 3311 GZ DORDRECHT, NETHERLANDS}
}

@article{pal2023optimizing,
  title={Optimizing the Last Mile: Advanced Predictive Analytics for Delivery Time Estimation in Supply Chain Logistics},
  author={Pal, Subharun},
  journal={International Journal for Innovative Research in Multidisciplinary Field},
  volume={9},
  number={11},
  pages={55--61},
  year={2023}
}

@inproceedings{mohsin2017iot,
  title={IoT based cold chain logistics monitoring},
  author={Mohsin, Afreen and Yellampalli, Siva S},
  booktitle={2017 IEEE International Conference on Power, Control, Signals and Instrumentation Engineering (ICPCSI)},
  pages={1971--1974},
  year={2017},
  organization={IEEE}
}

@article{zhang2022iot,
  title={IoT network for international trade cold chain logistics tracking based on kalman algorithm},
  author={Zhang, Chao and Wei, Wei},
  journal={Computational Intelligence and Neuroscience},
  volume={2022},
  number={1},
  pages={1608167},
  year={2022},
  publisher={Wiley Online Library}
}

@article{blackburn2009supply,
  title={Supply chain strategies for perishable products: the case of fresh produce},
  author={Blackburn, Joseph and Scudder, Gary},
  journal={Production and Operations Management},
  volume={18},
  number={2},
  pages={129--137},
  year={2009},
  publisher={SAGE Publications Sage CA: Los Angeles, CA}
}

@misc{li2014sustainable,
  title={Sustainable food supply chain management},
  author={Li, Dong and Wang, Xiaojun and Chan, Hing Kai and Manzini, Riccardo},
  journal={International Journal of Production Economics},
  volume={152},
  pages={1--8},
  year={2014},
  publisher={Elsevier}
}

@article{folkerts1997challenges,
  title={Challenges in international food supply chains: vertical co-ordination in the European agribusiness and food industries},
  author={Folkerts, Henk and Koehorst, Hans},
  journal={Supply Chain Management: An International Journal},
  volume={2},
  number={1},
  pages={11--14},
  year={1997},
  publisher={MCB UP Ltd}
}

@article{abolghasemi2020demand,
  title={Demand forecasting in supply chain: The impact of demand volatility in the presence of promotion},
  author={Abolghasemi, Mahdi and Beh, Eric and Tarr, Garth and Gerlach, Richard},
  journal={Computers \& Industrial Engineering},
  volume={142},
  pages={106380},
  year={2020},
  publisher={Elsevier}
}

@article{haberleitner2010implementation,
  title={Implementation of a demand planning system using advance order information},
  author={Haberleitner, Helmut and Meyr, Herbert and Taudes, Alfred},
  journal={International Journal of Production Economics},
  volume={128},
  number={2},
  pages={518--526},
  year={2010},
  publisher={Elsevier}
}

@article{kumar2020big,
  title={A big data driven framework for demand-driven forecasting with effects of marketing-mix variables},
  author={Kumar, Ajay and Shankar, Ravi and Aljohani, Naif Radi},
  journal={Industrial marketing management},
  volume={90},
  pages={493--507},
  year={2020},
  publisher={Elsevier}
}

@article{moroff2021machine,
  title={Machine Learning and statistics: A Study for assessing innovative demand forecasting models},
  author={Moroff, Nikolas Ulrich and Kurt, Ersin and Kamphues, Josef},
  journal={Procedia Computer Science},
  volume={180},
  pages={40--49},
  year={2021},
  publisher={Elsevier}
}

@article{abbasimehr2020optimized,
  title={An optimized model using LSTM network for demand forecasting},
  author={Abbasimehr, Hossein and Shabani, Mostafa and Yousefi, Mohsen},
  journal={Computers \& industrial engineering},
  volume={143},
  pages={106435},
  year={2020},
  publisher={Elsevier}
}

@article{khashei2011novel,
  title={A novel hybridization of artificial neural networks and ARIMA models for time series forecasting},
  author={Khashei, Mehdi and Bijari, Mehdi},
  journal={Applied soft computing},
  volume={11},
  number={2},
  pages={2664--2675},
  year={2011},
  publisher={Elsevier}
}

@article{min2010artificial,
  title={Artificial intelligence in supply chain management: theory and applications},
  author={Min, Hokey},
  journal={International Journal of Logistics: Research and Applications},
  volume={13},
  number={1},
  pages={13--39},
  year={2010},
  publisher={Taylor \& Francis}
}

@article{szul2020neural,
  title={Neural methods comparison for prediction of heating energy based on few hundreds enhanced buildings in four season’s climate},
  author={Szul, Tomasz and N{\k{e}}cka, Krzysztof and Mathia, Thomas G},
  journal={Energies},
  volume={13},
  number={20},
  pages={5453},
  year={2020},
  publisher={MDPI}
}

@article{catal2015improvement,
  title={Improvement of demand forecasting models with special days},
  author={Catal, Cagatay and Fenerci, Ayse and Ozdemir, Burcak and Gulmez, Onur},
  journal={Procedia Computer Science},
  volume={59},
  pages={262--267},
  year={2015},
  publisher={Elsevier}
}

@article{shahrabi2013developing,
  title={Developing a hybrid intelligent model for forecasting problems: Case study of tourism demand time series},
  author={Shahrabi, Jamal and Hadavandi, Esmaeil and Asadi, Shahrokh},
  journal={Knowledge-Based Systems},
  volume={43},
  pages={112--122},
  year={2013},
  publisher={Elsevier}
}

@article{potovcnik2019comparison,
  title={A comparison of models for forecasting the residential natural gas demand of an urban area},
  author={Poto{\v{c}}nik, Primo{\v{z}} and {\v{S}}ilc, Jurij and Papa, Gregor and others},
  journal={Energy},
  volume={167},
  pages={511--522},
  year={2019},
  publisher={Elsevier}
}

@article{amirkolaii2017demand,
  title={Demand forecasting for irregular demands in business aircraft spare parts supply chains by using artificial intelligence (AI)},
  author={Amirkolaii, K Nemati and Baboli, Armand and Shahzad, MK and Tonadre, R},
  journal={IFAC-PapersOnLine},
  volume={50},
  number={1},
  pages={15221--15226},
  year={2017},
  publisher={Elsevier}
}

@article{aktepe2021demand,
  title={Demand forecasting application with regression and artificial intelligence methods in a construction machinery company},
  author={Aktepe, Adnan and Yan{\i}k, Emre and Ers{\"o}z, S{\"u}leyman},
  journal={Journal of Intelligent Manufacturing},
  volume={32},
  number={6},
  pages={1587--1604},
  year={2021},
  publisher={Springer}
}

@article{rahman2018predicting,
  title={Predicting electricity consumption for commercial and residential buildings using deep recurrent neural networks},
  author={Rahman, Aowabin and Srikumar, Vivek and Smith, Amanda D},
  journal={Applied energy},
  volume={212},
  pages={372--385},
  year={2018},
  publisher={Elsevier}
}

@article{hochreiter1997long,
  title={Long short-term memory},
  author={Hochreiter, Sepp and Schmidhuber, J{\"u}rgen},
  journal={Neural computation},
  volume={9},
  number={8},
  pages={1735--1780},
  year={1997},
  publisher={MIT press}
}

@article{wu2018remaining,
  title={Remaining useful life estimation of engineered systems using vanilla LSTM neural networks},
  author={Wu, Yuting and Yuan, Mei and Dong, Shaopeng and Lin, Li and Liu, Yingqi},
  journal={Neurocomputing},
  volume={275},
  pages={167--179},
  year={2018},
  publisher={Elsevier}
}

@article{siami2019comparative,
  title={A comparative analysis of forecasting financial time series using arima, lstm, and bilstm},
  author={Siami-Namini, Sima and Tavakoli, Neda and Namin, Akbar Siami},
  journal={arXiv preprint arXiv:1911.09512},
  year={2019}
}

@article{ramirez2021neural,
  title={Neural Multi-Quantile Forecasting for Optimal Inventory Management},
  author={Ram{\'\i}rez, Federico Garza},
  journal={arXiv preprint arXiv:2112.05673},
  year={2021}
}

@inproceedings{sak2014long,
  title={Long short-term memory recurrent neural network architectures for large scale acoustic modeling.},
  author={Sak, Hasim and Senior, Andrew W and Beaufays, Fran{\c{c}}oise and others},
  booktitle={Interspeech},
  volume={2014},
  pages={338--342},
  year={2014}
}

@article{liu2022effect,
  title={Effect of weather on online food ordering},
  author={Liu, Da and Wang, Wenbo and Zhao, Yinchuan},
  journal={Kybernetes},
  volume={51},
  number={1},
  pages={165--209},
  year={2022},
  publisher={Emerald Publishing Limited}
}

@article{yao2023weather,
  title={Weather and time factors impact on online food delivery sales: a comparative analysis of three Chinese cities},
  author={Yao, Wang and Zhao, Hongying and Liu, Luning},
  journal={Theoretical and Applied Climatology},
  volume={153},
  number={3},
  pages={1425--1438},
  year={2023},
  publisher={Springer}
}

@article{negi2019transportation,
  title={Transportation lead time in perishable food value chains: an Indian perspective},
  author={Negi, Saurav and Wood, Lincoln C},
  journal={International Journal of Value Chain Management},
  volume={10},
  number={4},
  pages={290--315},
  year={2019},
  publisher={Inderscience Publishers (IEL)}
}

@article{andreyeva2010impact,
  title={The impact of food prices on consumption: a systematic review of research on the price elasticity of demand for food},
  author={Andreyeva, Tatiana and Long, Michael W and Brownell, Kelly D},
  journal={American journal of public health},
  volume={100},
  number={2},
  pages={216--222},
  year={2010},
  publisher={American Public Health Association}
}

@article{nair2011supply,
  title={Supply network topology and robustness against disruptions--an investigation using multi-agent model},
  author={Nair, Anand and Vidal, Jos{\'e} M},
  journal={International Journal of Production Research},
  volume={49},
  number={5},
  pages={1391--1404},
  year={2011},
  publisher={Taylor \& Francis}
}

@article{zhang2003time,
  title={Time series forecasting using a hybrid ARIMA and neural network model},
  author={Zhang, G Peter},
  journal={Neurocomputing},
  volume={50},
  pages={159--175},
  year={2003},
  publisher={Elsevier}
}

@article{syntetos2005accuracy,
  title={The accuracy of intermittent demand estimates},
  author={Syntetos, Aris A and Boylan, John E},
  journal={International Journal of forecasting},
  volume={21},
  number={2},
  pages={303--314},
  year={2005},
  publisher={Elsevier}
}

@article{chatfield2004bullwhip,
  title={The bullwhip effect—impact of stochastic lead time, information quality, and information sharing: a simulation study},
  author={Chatfield, Dean C and Kim, Jeon G and Harrison, Terry P and Hayya, Jack C},
  journal={Production and operations management},
  volume={13},
  number={4},
  pages={340--353},
  year={2004},
  publisher={Wiley Online Library}
}

@article{wangphanich2010analysis,
  title={Analysis of the bullwhip effect in multi-product, multi-stage supply chain systems--a simulation approach},
  author={Wangphanich, Pilada and Kara, Sami and Kayis, Berman},
  journal={International journal of production Research},
  volume={48},
  number={15},
  pages={4501--4517},
  year={2010},
  publisher={Taylor \& Francis}
}

@article{tkachuk2025consumer,
  title={Consumer Transactions Simulation Through Generative Adversarial Networks Under Stock Constraints in Large-Scale Retail},
  author={Tkachuk, Sergiy and {\L}ukasik, Szymon and Wr{\'o}blewska, Anna},
  journal={Electronics},
  volume={14},
  number={2},
  pages={284},
  year={2025},
  publisher={MDPI}
}

@incollection{tjen2023demand,
  title={Demand Forecasting Using Time Series and ANN with Inventory Control to Reduce Bullwhip Effect on Home Appliances Electronics Distributors},
  author={Tjen, Stiven and Gozali, Lina and Kristina, Helena Juliana and Gunadi, Ariawan and Irawan, Agustinus Purna},
  booktitle={Mobile Computing and Sustainable Informatics: Proceedings of ICMCSI 2023},
  pages={285--300},
  year={2023},
  publisher={Springer}
}

@article{udenio2023exponential,
  title={Exponential smoothing forecasts: taming the bullwhip effect when demand is seasonal},
  author={Udenio, Maximiliano and Vatamidou, Eleni and Fransoo, Jan C},
  journal={International Journal of Production Research},
  volume={61},
  number={6},
  pages={1796--1813},
  year={2023},
  publisher={Taylor \& Francis}
}

@article{chen2024inventory,
  title={The inventory bullwhip effect in the online retail supply chain considering the price discount based on different forecasting methods},
  author={Chen, Qiang and Zhang, Xiaoqing and Liu, Shuang and Zhu, Weixiao},
  journal={The Engineering Economist},
  volume={69},
  number={2},
  pages={108--128},
  year={2024},
  publisher={Taylor \& Francis}
}

@inproceedings{viloria2019demand,
  title={Demand Forecasting Method Using Artificial Neural Networks},
  author={Viloria, Amelec and Matos, Luisa Fernanda Arrieta and Gait{\'a}n, Mercedes and Palma, Hugo Hern{\'a}ndez and Guzm{\'a}n, Yasmin Fl{\'o}rez and V{\'a}squez, Luis Cabas and Mercado, Carlos Vargas and Lezama, Omar Bonerge Pineda},
  booktitle={Dependability in Sensor, Cloud, and Big Data Systems and Applications: 5th International Conference, DependSys 2019, Guangzhou, China, November 12--15, 2019, Proceedings 5},
  pages={383--391},
  year={2019},
  organization={Springer}
}

@article{alabdulkarim2020minimizing,
  title={Minimizing the bullwhip effect in a supply chain: a simulation approach using the beer game},
  author={Alabdulkarim, Abdullah A},
  journal={Simulation},
  volume={96},
  number={9},
  pages={737--752},
  year={2020},
  publisher={SAGE Publications Sage UK: London, England}
}

@article{chiadamrong2021meta,
  title={Meta-prediction models for bullwhip effect prediction of a supply chain using regression analysis},
  author={Chiadamrong, Navee and Sarnrak, Nont},
  journal={International Journal of Information Systems and Supply Chain Management (IJISSCM)},
  volume={14},
  number={4},
  pages={36--71},
  year={2021},
  publisher={IGI Global}
}

@article{rezaeefard2024present,
  title={Present a mixed approach of neural network and bat algorithm to predict customer demand in the supply chain to reduce the Bullwhip effect},
  author={Rezaeefard, Milad and Pilevari, Nazanin and Faezy Razi, Farshad},
  journal={International Journal of Nonlinear Analysis and Applications},
  volume={15},
  number={4},
  pages={65--78},
  year={2024},
  publisher={Semnan University}
}

@phdthesis{kleinemolen2024inventory,
  title={Inventory Optimization and Simulation Analysis for Supply Chain Disruption Events},
  author={Kleinemolen, Ian},
  year={2024},
  school={Massachusetts Institute of Technology}
}

@article{yang2022supply,
  title={Supply chain information collaborative simulation model integrating multi-agent and system dynamics},
  author={Yang, Ning and Ding, Yingzi and Leng, Junge and Zhang, Lei},
  journal={Promet-Traffic\&Transportation},
  volume={34},
  number={5},
  pages={711--724},
  year={2022},
  publisher={Sveu{\v{c}}ili{\v{s}}te u Zagrebu Fakultet prometnih znanosti}
}

@book{chopra2007supply,
  title={Supply chain management. Strategy, planning \& operation},
  author={Chopra, Sunil and Meindl, Peter},
  year={2007},
  publisher={Springer}
}

@article{thecharmsofs2001og,
  title={Log-no rmal Distributions across the Sciences: Keys and Clues},
  author={Thecharmsofs, ON and BILITY—NORMAL, VA RIABILITYANDP RO BA and LOG, OR and AT, TH},
  journal={BioScience},
  volume={51},
  number={5},
  year={2001}
}

@book{christopher2022logistics,
  title={Logistics and supply chain management},
  author={Christopher, Martin},
  year={2022},
  publisher={Pearson Uk}
}

@article{hadley1963analysis,
  title={Analysis of inventory systems},
  author={Hadley, George and Whitin, Thomson M},
  journal={(No Title)},
  year={1963}
}

@article{feizabadi2022machine,
  title={Machine learning demand forecasting and supply chain performance},
  author={Feizabadi, Javad},
  journal={International Journal of Logistics Research and Applications},
  volume={25},
  number={2},
  pages={119--142},
  year={2022},
  publisher={Taylor \& Francis}
}

@phdthesis{paruthipattu2021demand,
  title={Demand Forecasting based on External Factors using Clustering and Machine learning},
  author={Paruthipattu, Sruthi Prabakaran},
  year={2021},
  school={Dublin, National College of Ireland}
}

\end{document}